\documentclass[conference]{IEEEtran}
\IEEEoverridecommandlockouts
\usepackage{cite}
\usepackage{amsmath,amssymb,amsfonts}
\usepackage{graphicx}
\usepackage{textcomp}
\usepackage{xcolor}
\def\BibTeX{{\rm B\kern-.05em{\sc i\kern-.025em b}\kern-.08em
    T\kern-.1667em\lower.7ex\hbox{E}\kern-.125emX}}
\usepackage{times}
\usepackage{latexsym}
\usepackage[T1]{fontenc}
\usepackage[utf8]{inputenc}
\usepackage{microtype}

\usepackage{amsmath}
\usepackage{adjustbox}     
\usepackage{hyperref}
\usepackage{microtype}
\usepackage{graphicx}
\usepackage{amsmath}
\usepackage{lipsum}
\usepackage{algpseudocode}
\usepackage{amsfonts}
\usepackage{subcaption}
\usepackage{comment}
\usepackage{url}
\usepackage{multirow}
\usepackage{multirow}
\usepackage{booktabs}
\DeclareUnicodeCharacter{2212}{-}
\usepackage{makecell}

\usepackage{float}

\begin{document}
\title{Understanding COVID-19 Vaccine Campaign on Facebook using Minimal Supervision}

\author{\IEEEauthorblockN{Tunazzina Islam}
\IEEEauthorblockA{\textit{Department of Computer Science} \\
\textit{Purdue University, West Lafayette, IN 47907, USA}\\
islam32@purdue.edu}
\and
\IEEEauthorblockN{Dan Goldwasser}
\IEEEauthorblockA{\textit{Department of Computer Science} \\
\textit{Purdue University, West Lafayette, IN 47907, USA }\\
dgoldwas@purdue.edu}
}

\maketitle

\begin{abstract}
In the age of social media, where billions of internet users share information and opinions, the negative impact of pandemics is not limited to the physical world. It provokes a surge of incomplete, biased, and incorrect information, also known as an infodemic. This global infodemic jeopardizes measures to control the pandemic by creating panic, vaccine hesitancy, and fragmented social response. Platforms like Facebook allow advertisers to adapt their messaging to target different demographics and help alleviate or exacerbate the infodemic problem depending on their content. In this paper, we propose a minimally supervised multi-task learning framework for understanding messaging on Facebook related to the COVID vaccine by identifying ad themes and moral foundations. Furthermore, we perform a more nuanced thematic analysis of messaging tactics of vaccine campaigns on social media so that policymakers can make better decisions on pandemic control.
\end{abstract}

\section{Introduction}
Since January 2020, worldwide public health has been threatened by the novel coronavirus  -- the outbreak was declared a global pandemic by the World Health Organization (WHO) \cite{cucinotta2020declares}. COVID-19 is the first pandemic in the history in which technology and social media are being used on a massive scale to keep people safe, informed, productive, and connected. Yet, at the same time, the growing proliferation of social media can be used for spreading hoaxes and false information leading to what is commonly referred to as an infodemic \cite{tagliabue2020pandemic}. 
\begin{figure}[htbp]
  \centering  
  \includegraphics[width= .5 \textwidth]{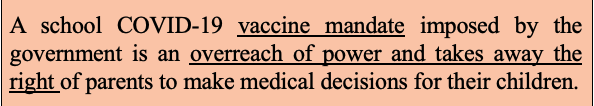}
    \caption{Example of an ad highlighting analysis dimensions (1) theme, (2) moral foundation.}
    \label{fig:ad}
\end{figure}
%
\begin{table}[t]
\begin{center}
 \scalebox{.75}{\begin{tabular}{>{\arraybackslash}m{10cm}}
 \toprule
\textsc{\textbf{Care/Harm:}} Saying that someone other than the speaker deserves care or gets harmed. Reflects the base of Maslow’s Hierarchy of
Needs \cite{mcleod2007maslow}. Security, Shelter, Food, Water, Warmth. \\
\hline
\textsc{\textbf{Fairness/Cheating:}} Justice, rights, and autonomy, comparison to other groups. Equality of Opportunities. Social Intolerance to ``Free-Rider".\\
\hline
\textsc{\textbf{Loyalty/Betrayal:}} Patriotism and self-sacrifice for the group (Or failure to provide it, in the case of betrayal). It is active anytime people feel that it’s “one for all, and all for one.”\\
\hline
\textsc{\textbf{Authority/Subversion:}} Deference (or opposition) to legitimate authority and respect for traditions. Social order and the obligations of hierarchical relationships, such as obedience, respect, and the fulfillment of role-based duties. \\
\hline
\textsc{\textbf{Sanctity/Degradation:}} Not simply a religious value. Respect for the human spirit. Social aversion of personal degradation. Degradation causes disgust. Physical and spiritual contagion, including virtues of chastity, wholesomeness, and control of desires. It underlies the widespread idea that the body is a temple which can be desecrated by immoral activities and contaminants (an idea not unique to religious traditions).\\
\hline 
\textsc{\textbf{Liberty/Oppression:}} Feelings of people towards those who dominate them and restrict their liberty. The hatred of bullies and dominators motivates people to come together, in solidarity, to oppose or take down the oppressor.\\
\hline 
\textsc{\textbf{None:}}  Does not fall under any other foundation.\\
\bottomrule
\end{tabular}}
\caption[]{Six basic moral foundations.\footnotemark}
\label{tab:mf}
\end{center}
\end{table}
\footnotetext{\href{https://www.networkforphl.org/wp-content/uploads/2020/04/COVID-19-Guidance-Framing-Coronavirus-Messaging-Using-the-Moral-Foundations-Theory-Framework.pdf}{networkforphl.org}}
Social media discourse help increase polarization around topics related to COVID-19 vaccines, such as the vaccine mandate, natural immunity, vaccine efficacy, religious sentiment, and vaccine equity. Moral Foundation Theory (MFT) \cite{haidt2004intuitive,haidt2007morality} suggests a theoretical framework for analyzing the morality, containing six basic moral foundations (Table. \ref{tab:mf}). Past work has shown that the theory can help explain ideological differences and social group membership~\cite{graham2009liberals,graham2013moral,weber2013moral,silver2017conservatives,johnson2018classification,roy2021identifying,turner2021conservatives}.
Often there is a significant correlation between the vaccine debate and its moral foundation (MF).

However, sponsored contents have been used to reach more people on social media to disseminate their agendas.
For example, Fig. \ref{fig:ad} presents a sponsored ad on Facebook representing `vaccine mandate' as the ad theme and \textit{overreach of power and takes away the right}, falling under the `liberty/oppression' moral foundation.
Therefore, detecting moral foundation and theme from text are vital components in understanding advertisers' intention, key talking points, policies. 


Our goal in this paper is to take a first step toward analyzing the landscape of vaccination campaigns on social media. We focus our experiments on a timely topic, COVID-19 vaccination campaign. Our main contributions are twofold: (1) to identify the ad theme \& moral foundation; (2) to build on this characterization to analyze the messaging across different demographics, geographic, and timelines. We analyze over $28K$ COVID vaccine related ads on Facebook, associating ads with $6$ moral foundations and including `none', it's a $7$-class classification problem. We also identify the theme of the ads, a $15$-class classification problem.

Our theme analysis is motivated by previous studies \cite{wawrzuta2021arguments,weinzierl2021misinformation,pacheco2022holistic} that created code-book of COVID-19 vaccine arguments. 
Besides, our moral foundation analysis is inspired by social science studies \cite{pagliaro2021trust,diaz2021reactance,chan2021moral} that demonstrated relation between moral foundation and COVID related health decisions. 
\begin{table}[]
\begin{adjustbox}{width=1\columnwidth,center}
    \begin{tabular}{ll}
      \toprule
      \textbf{Moral Foundations} & \textbf{Example of messages} \\
      \midrule
      \textbf{Care/Harm} & Protect yourself and others. \\
      ~ & Help those most vulnerable. \\
      ~ & Public health can assist you. \\
      ~ & Stay healthy and safe. \\
       
      \textbf{Fairness/Cheating} & Everyone has an interest in beating this outbreak. \\
      ~ & Infection does not discriminate. \\
      ~ & We have an interest in everyone getting appropriate care. \\
      ~ & Vaccine should be free for everyone. \\
      
      \textbf{Loyalty/Betrayal} &  Do your part, take the shot for your family, friends, country. \\
      ~ & We need to protect our community. \\
      ~ & I’m loyal to you and want to keep you safe. \\
      ~ & Limited resources should go first to healthcare workers and those caring for us. \\
      
      \textbf{Authority/Subversion} & Scientific evidence and common sense show that protective measures really work. \\
      ~ & Listen to your local public health official. \\
      ~ & Respect healthcare workers and the risks they are taking. \\
      ~ & Trust science. \\
      ~ & Be a good role model for others. \\
      
      \textbf{Sanctity/Degradation} & Be willing to sacrifice your wants for community needs.\\
      ~ & Help nurture the spirits of those needing comfort. \\
      ~ & Look for ways to serve others. \\
      
      \textbf{Liberty/Oppression} & COVID can threaten our safety and freedom. \\
      ~ & We want our community to be free from fear of contagion. \\
      ~ & The quicker we beat this, the quicker we recover and return to normal.\\
      \bottomrule
    \end{tabular}%
   \end{adjustbox}
   \caption{Example messages corresponding to each moral foundation provided to annotators.}
    \label{tab:mf_example}
\end{table}

In this paper, we suggest a minimally supervised multi-task learning approach to understand COVID-19 vaccine campaign in Facebook. 
The purpose of minimal supervision is to compensate for the lack of annotated data by exploiting the maximum potential of the available data. 
For MF, we generate weak labels from dedicated lexicons developed for identifying moral foundation. For theme, we use a pre-trained textual inference model to identify
paraphrases in a large collection of COVID-19 vaccination ads from Facebook and assign theme based on cluster assignment (Details in \ref{wk_label}). 
We focus on the following research questions (RQ) to analyze vaccine campaigns on social media:
 \begin{itemize}
    \item \textbf{RQ1.} What are the narratives of the messaging? (section \ref{narrative})
    \item \textbf{RQ2.} How does entity type fulfill messaging roles? (section \ref{entity})
    \item \textbf{RQ3.} Which demographics and geographic are reached by the advertisers and their messages? (section \ref{Demographic})
    \item \textbf{RQ4.} Do ads follow current COVID status? (section \ref{granger})
\end{itemize}
We summarize the main contributions of this paper as the following:
\begin{enumerate}
    \item We formulate a novel problem of using minimal supervision to analyze the landscape of vaccine campaigns on Facebook. Our dataset is publicly available here\footnote{\url{https://github.com/tunazislam/Covid_FB_AD_MinimalSup}}.
    \item We suggest a minimally supervised multi-task learning framework with three different learning strategies to identify ad theme and moral foundation.
    \item We investigate the COVID vaccine ads on Facebook from four angles: narratives (thematic and moral foundation analysis), entity types (who is funding the ad), reach (who saw the ads), and whether the ads reflect current COVID situations.
\end{enumerate}
\section{Related Work}
Recent studies have shown narrative analysis and opinion mining of COVID-19 pandemic discourse in social media and news media \cite{shurafa2020political,muric2021covid,thelwall2021covid,krawczyk2021quantifying,jing2021characterizing,prado2022moral,shi2021psycho}. Also, there are recent studies on online perceptions about COVID-19 vaccination related to public health measures \cite{bokemper2022testing,mermin2022s,nan2022public} and moral
foundations \cite{graham2020faith,heine2021using,tarry2022political,coelho2022left,kyung2022political}.  Nowadays, targeted online advertising is one of the main communication channels, allowing hyper-local sponsors to campaign during the pandemic. Sponsored content on social media can be shared with various narratives, including information and misinformation, to disseminate agendas targeting specific demographics and geographic. Mejova and Kalimeri \cite{mejova2020covid} analyzed a smaller set of COVID-19 related Facebook ads messaging by identifying advertisers and their targets. Silva and Benevenuto \cite{silva2021covid} monitored COVID related Facebook Ads in Brazil to identify misinformation. Our work takes a different approach to analyze COVID vaccine related Facebook ads by identifying themes and moral foundation that motivate sponsors. Our work falls under the broad scope of weak supervision \cite{belem2021weakly,mekala2020meta,ratner2018snorkel,islam2022twitter} and multi-task learning \cite{caruana1997multitask,liu2016recurrent,liu2019multi,lu2020multi,vandenhende2021multi}. 
\section{Dataset Details}
\label{dataset}
We collected approximately $28,000$ COVID vaccine related ads focusing on United States from December 2020 - January 2022 using Facebook Ad Library API\footnote{\url{https://www.facebook.com/ads/library/api}} with the search term `COVID-19 vaccine', `COVID vaccine', `vaccination', `vaccine', `coronavirus vaccine', `corona vaccine'. Our collected ads were written in English. For each ad, the API provides
ad ID, title, ad body, funding entity, spend, impressions, distribution over impressions broken down by gender (male, female, unknown), age ($7$ groups), location down to states in the USA. 
We have duplicate content among those collected ads because the same ad has been targeted to different regions and demographics with unique ad id. We have $9,920$ ads with different contents. 
\begin{table}
\begin{adjustbox}{width=1\columnwidth,center}
    \begin{tabular}{ll}
      \toprule
      \textbf{Themes} & \textbf{Definition} \\
      \midrule
      \textbf{EncourageVaccination} & Promoting vaccination to control pandemic. \\
      \textbf{VaccineMandate} & Arguments about vaccine mandate, vaccine passport/card. \\
      \textbf{VaccineEquity} & Acknowledging no nation, state, or individual’s life \\
      ~ & is more important or more deserving than another’s. \\
      \textbf{VaccineEfficacy} & Arguments saying that the vaccine is safe, lessens the symptoms.  \\
    \textbf{GovDistrust} & Arguments saying people do not have trust on Governmental \\
     ~ & institutions or authority figures.\\
    \textbf{GovTrust} & Arguments saying people have trust on Governmental \\
     ~ & institutions or authority figures. \\
    \textbf{VaccineRollout} & Information about vaccination sites and availability of appointments.\\
    \textbf{VaccineSymptom} & Symptoms associated with the vaccine, e.g., fever, sore arm etc.\\
    \textbf{VaccineStatus} & Information regarding rate of vaccination, hospitalization, death etc.\\
    \textbf{VaccineReligion} & Arguments about religion and vaccine. \\
    \textbf{VaccineDevelopment} & Broadcasting information about the vaccine development and approval.\\
    \textbf{CovidPlan} & Good policies to deal with COVID-19. \\
    \textbf{VaccineMisinformation} & Conspiracy theories, fake news related to vaccine. \\
    \textbf{NaturalImmunity} & Natural methods of protection against COVID. \\
    \textbf{Vote} & Encourage residents to vote by iterating messages related to COVID vaccine.\\
   \bottomrule
    \end{tabular}%
    \end{adjustbox}
        \caption{Theme definition provided to the annotators.}
    \label{tab:theme}
\end{table}
\subsection{Data Annotation }
We manually annotated $557$ ads for themes and moral foundation. To ensure quality work, we provided annotators with $23$ examples covering all six moral foundations (Table. \ref{tab:mf_example}) and theme definition of $15$ themes (Table \ref{tab:theme}).
Two annotators from the Computer Science department manually annotated a subset of ads ($20\%$) to calculate inter-annotator agreement using Cohen’s Kappa coefficient \cite{cohen1960coefficient}. This subset has inter-annotator agreements of $73.80\%$ for MF and $65.60\%$ for theme which are substantial agreements. In case of a disagreement, we resolved it by discussion. Rest $80\%$ of the data was annotated by one of the graduate students between the two. We had one female and one male annotator, and the age range was $30-40$.
\section{Methodology}
\begin{figure}[htbp]
  \centering  
  \includegraphics[width= \columnwidth]{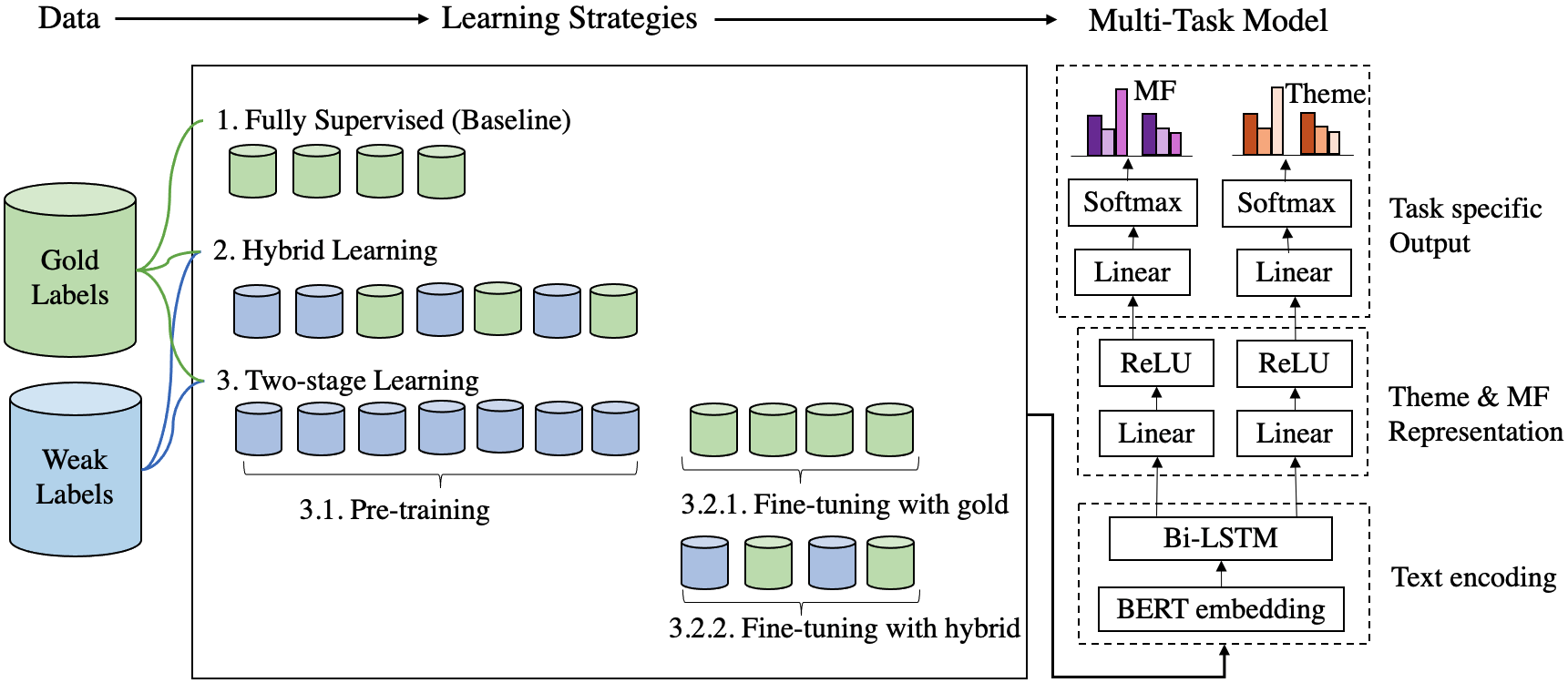}
\caption{An overview of our proposed framework. Best viewed in electronic format (zoomed in).}
    \label{fig:mtl}
\end{figure}

In this section, we start by describing the labeling technique that produces numerous but imprecise (weak) labels. Then, we put
forward two learning strategies to better exploit the available labels. Finally, we show the main components of multi-task learning model. An overview of the model is illustrated in Fig. \ref{fig:mtl}.
\subsection{Weak Label Generation}
\label{wk_label}
In this section, we describe how to generate weak labels from ad content that can be incorporated as weak sources in our model. 
\subsubsection{Themes}
\begin{figure}
\begin{subfigure}{.5\columnwidth}
  \centering
  \includegraphics[width=\textwidth]{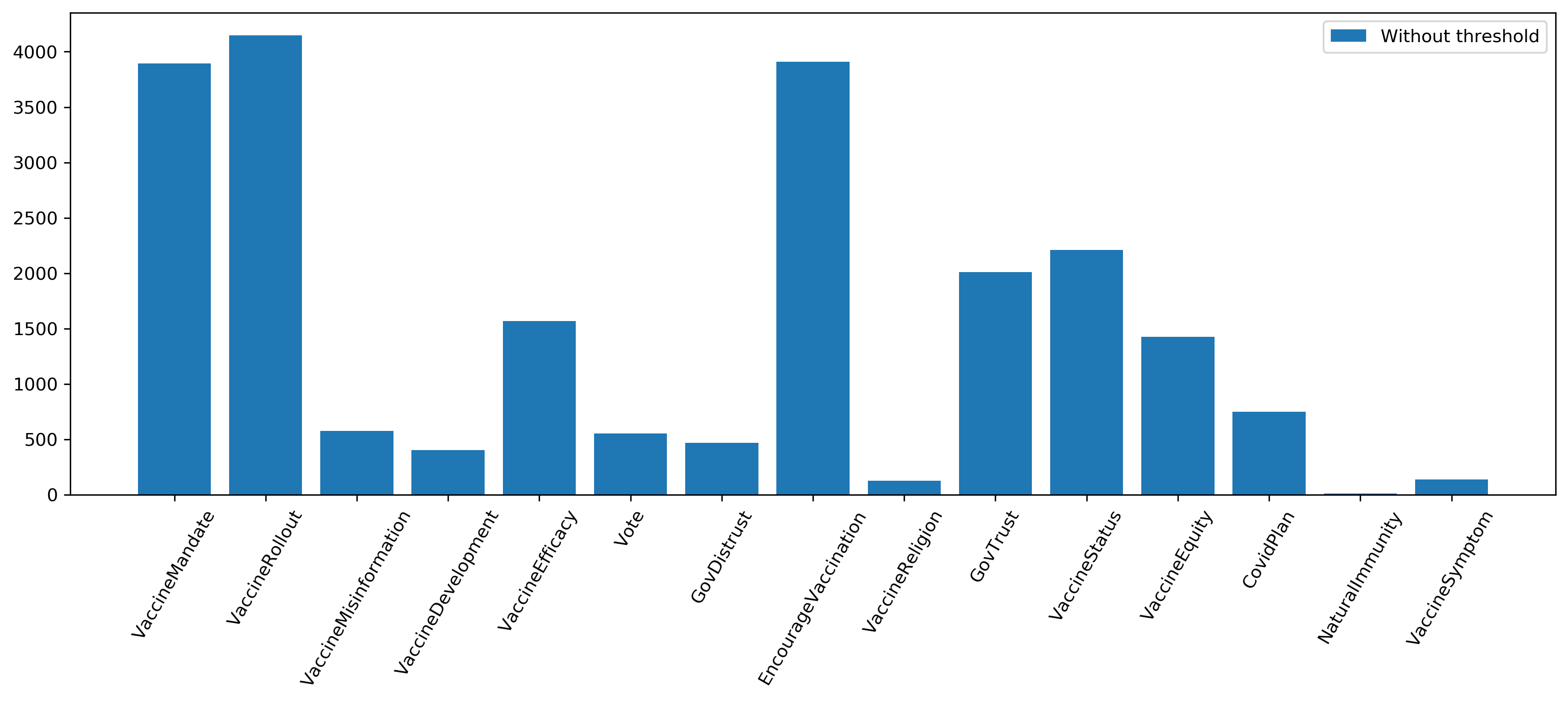}
  \caption{Without threshold. }
  \label{fig:Without_threshold}
\end{subfigure}%
\begin{subfigure}{.5\columnwidth}
  \centering
  \includegraphics[width=\textwidth]{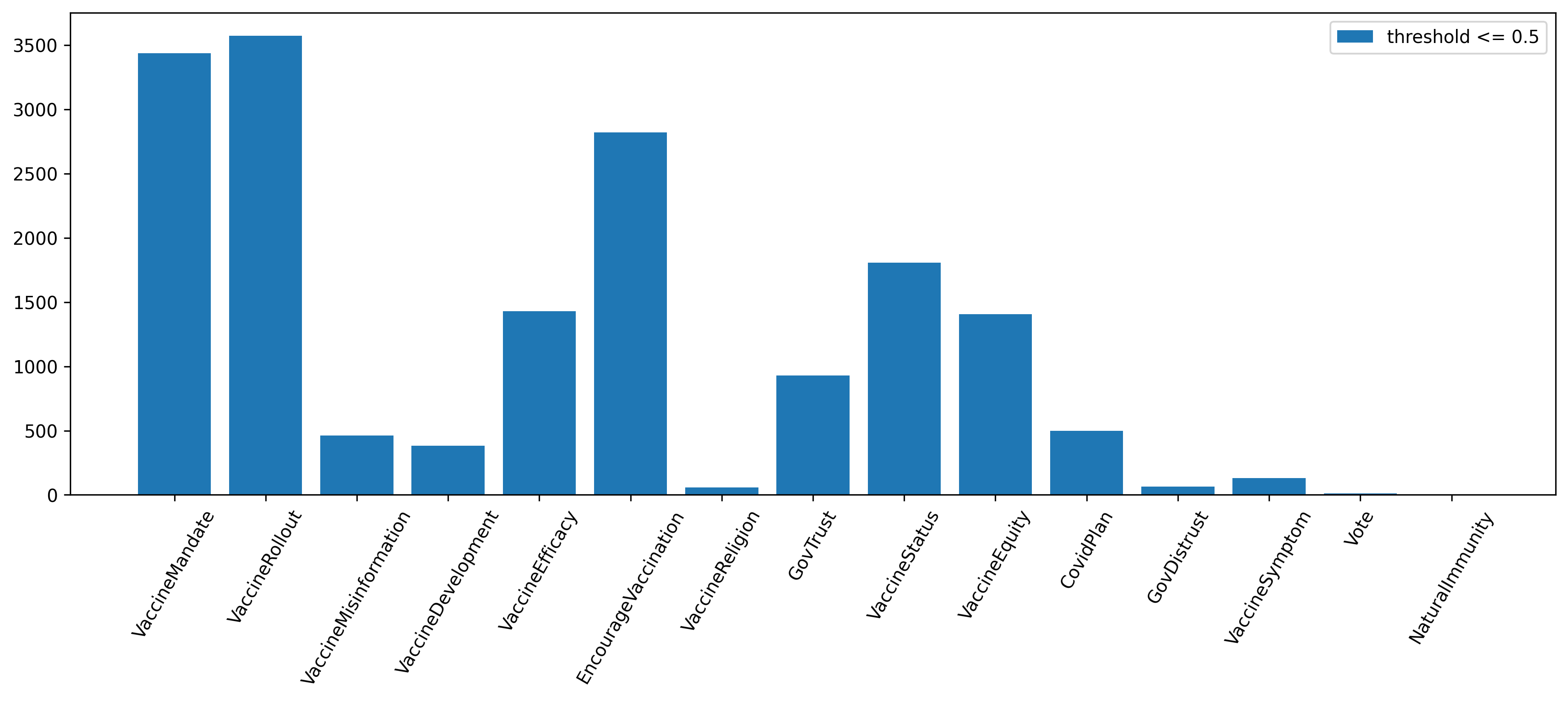}
  \caption{$threshold \leq 0.5 $}
  \label{fig:thr_0.5}
\end{subfigure}
\begin{subfigure}{.5\columnwidth}
  \centering
  \includegraphics[width=\textwidth]{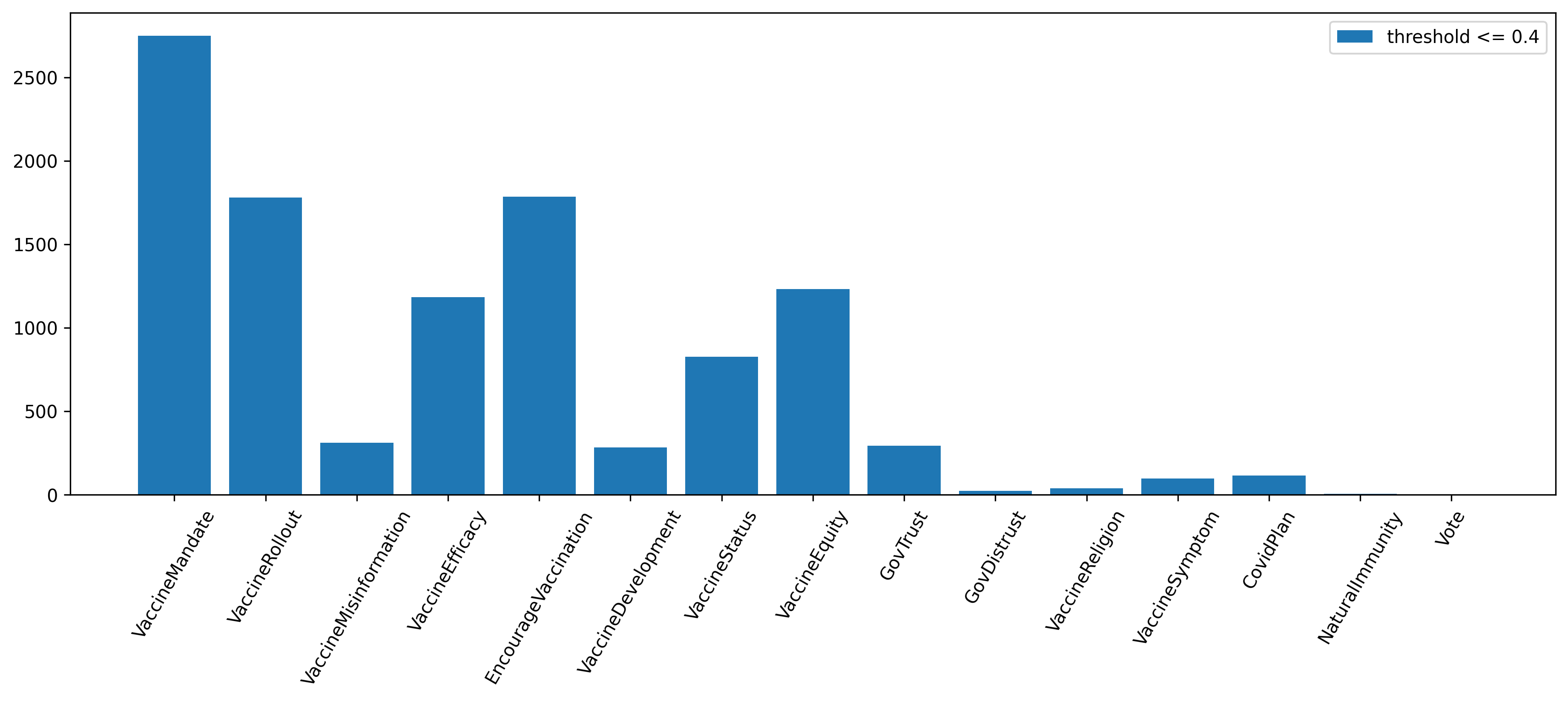}
  \caption{$threshold \leq 0.4 $ }
  \label{fig:thr_0.4}
\end{subfigure}%
\begin{subfigure}{.5\columnwidth}
  \centering
  \includegraphics[width=\textwidth]{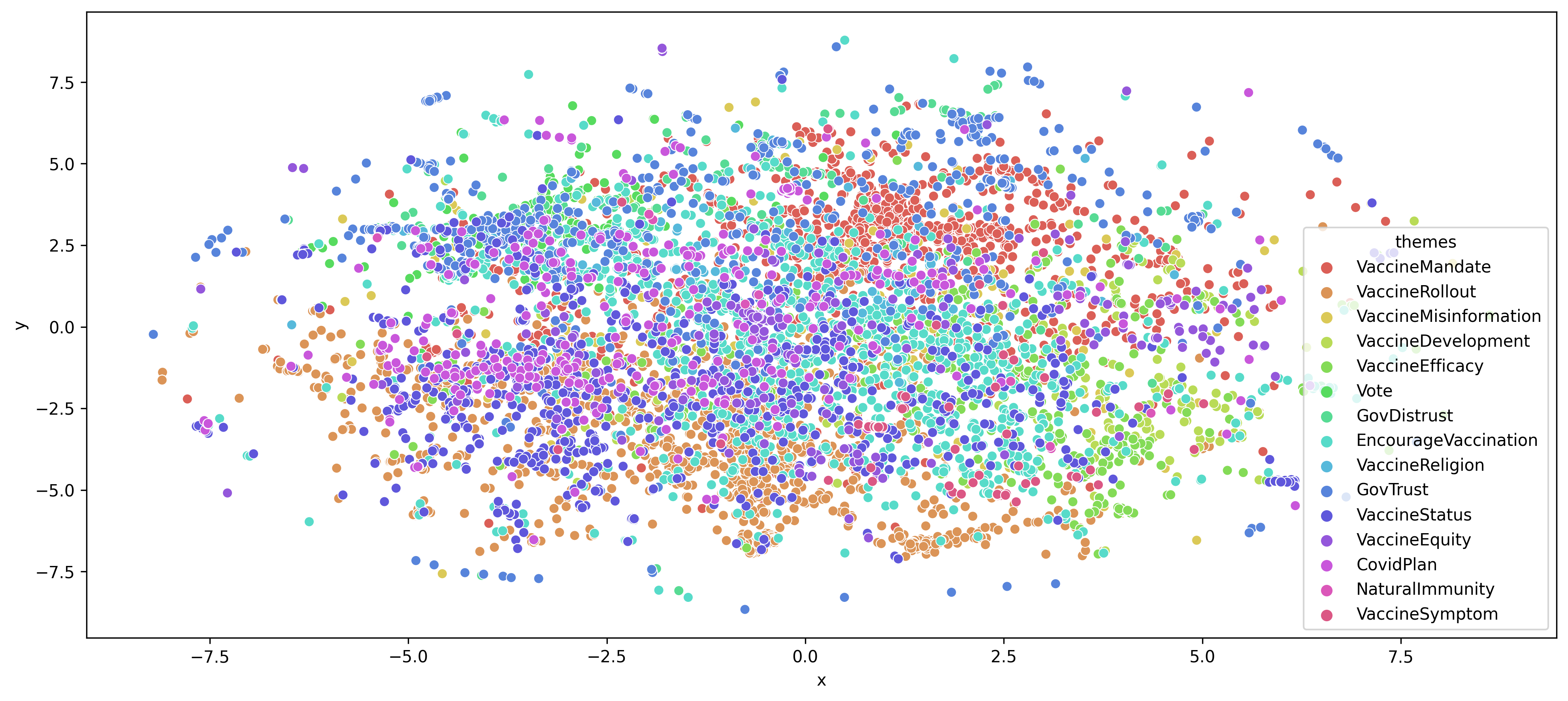}
  \caption{t-SNE without threshold. }
  \label{fig:tsne_without_threshold}
\end{subfigure}
\begin{subfigure}{.5\columnwidth}
  \centering
  \includegraphics[width=\textwidth]{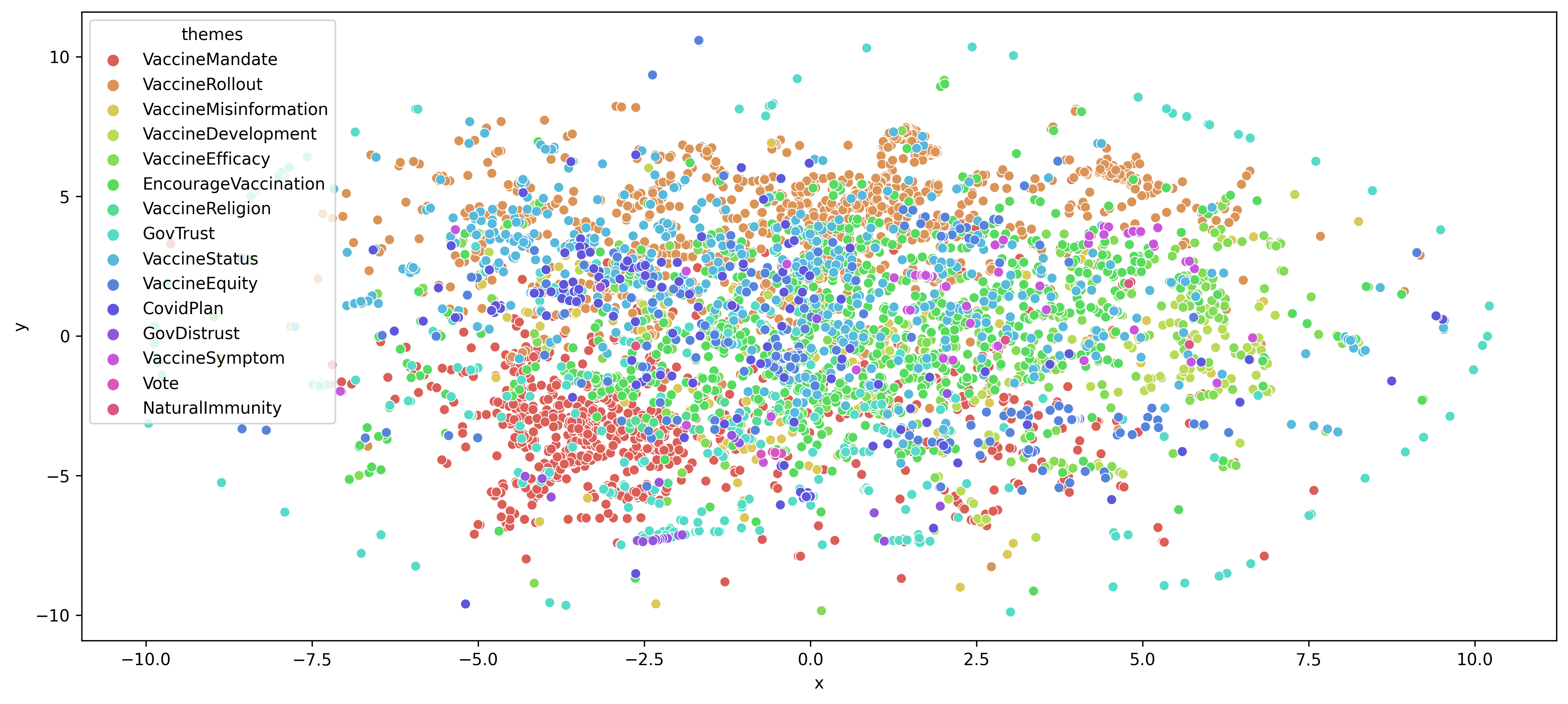}
  \caption{t-SNE ($threshold \leq 0.5 $)}
  \label{fig:tsne_thr_0.5}
\end{subfigure}%
\begin{subfigure}{.5\columnwidth}
  \centering
  \includegraphics[width=\textwidth]{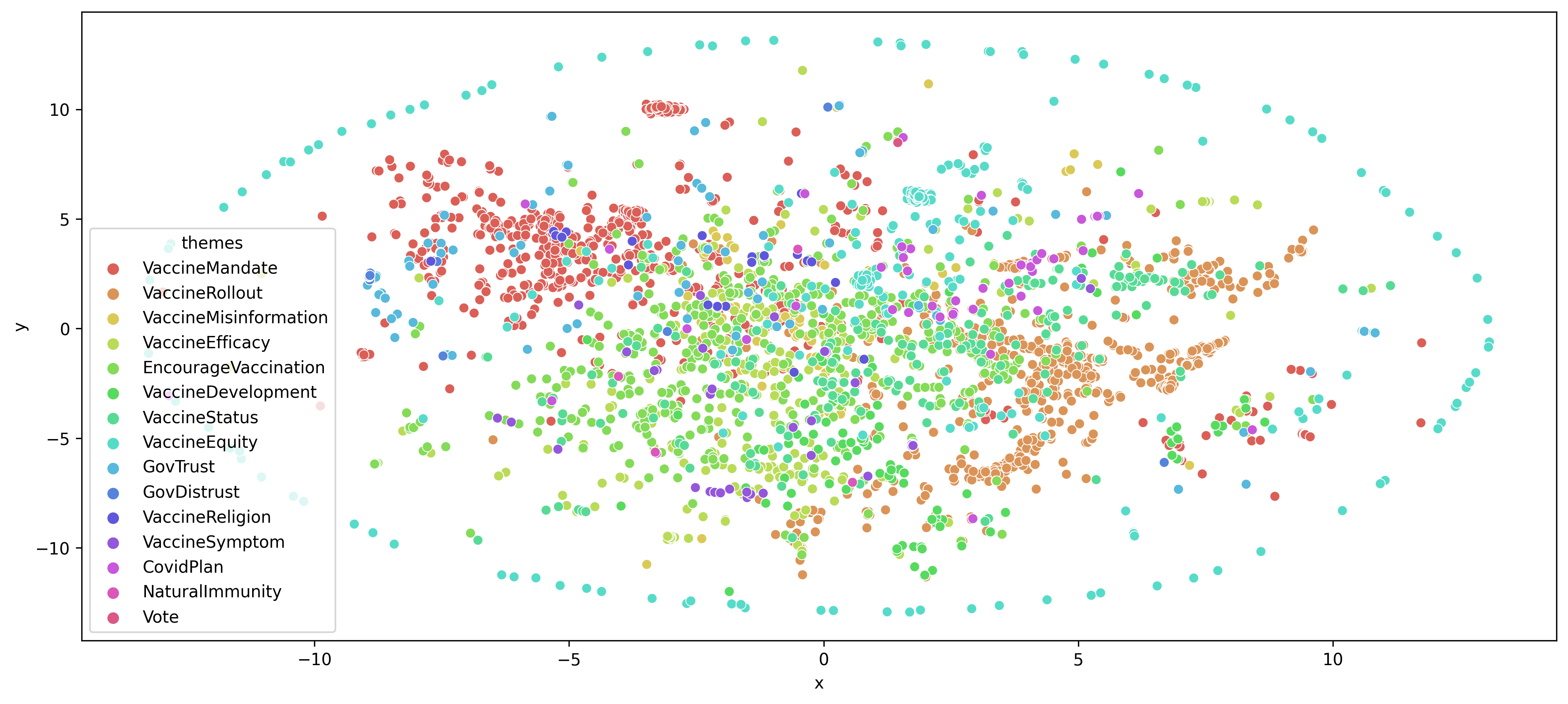}
  \caption{t-SNE ($threshold \leq 0.4 $)}
  \label{fig:tsne_thr_0.4}
\end{subfigure}
\caption{Bar plot and 2D visualization of cluster assignment for themes without and with threshold. Best viewed in electronic format (zoomed in).}
\label{fig:bar_tsne}
\end{figure}

First, we go through relevant research conducted by the health informatics, computational social science, public health, psychology communities 
and ground our analysis through constructing a
list of potential themes for COVID related ads \cite{pacheco2022holistic,weinzierl2022hesitancy,cheng2022debate,ehde2021covid,roozenbeek2020susceptibility}. Then, we consult with two researchers in Computational Social Science and finalize the relevant themes with corresponding phrases. The full list of phrases for each theme can be observed in Table \ref{tab:thm_phrs}.
To generate the weak labels for themes, we ground the phrases (from corresponding theme) in a set of $28k$ unlabelled COVID-19 vaccine related ads and match similarity between their Sentence BERT embeddings \cite{reimers2019sentence}. We measure the cluster purity using silhoutte score \cite{rousseeuw1987silhouettes}. We use  threshold based on closest distance to limit assignments. 
Bar plot with the number of assigned ads to each cluster with and without threshold and $2D$ visualizations of clusters are shown using t-SNE \cite{vanDerMaaten2008} in Fig. \ref{fig:bar_tsne}.
We use $threshold \leq 0.5 $ resulting $21,851$ ads.
\subsubsection{Moral Foundation}
Weak label for moral foundation is generated by analyzing text based on the MFT
relying on the use of a lexical resource,
the Moral Foundations Dictionary (MFD) \cite{graham2009liberals}. 
Similar to Linguistic Inquiry and Word Count (LIWC) \cite{pennebaker2001linguistic,tausczik2010psychological}, MFD associates a list of related words with each one of the moral foundations. 
We analyze ad's text by
counting the number of occurrences of words in
the text which also match the words in the MFD. In this process, same ad might get multiple moral foundations based on lexicon matching. To assign one MF for each ad, we pick the MF having the highest number of keyword matching with our text. Given that MFD does not have lexicon for liberty/oppression moral foundation, we use the same lexicon curated by Pacheco et al. \cite{pacheco2022holistic}. We annotate an ad as liberty/oppression MF if it contains at least two keywords.
\subsubsection{Quality of Weak Label}
To assess the weak label quality, we compare the weak labels with the ground truth labels ($557$ ads). The accuracy and macro-avg F1 score of the weak label for theme are $0.513$ and $0.337$ respectively. For moral foundation, the accuracy and macro-avg F1 score are $0.417$ and $0.248$ correspondingly. We observe that the accuracy and macro-avg F1 score of the weak label are significantly better than random ($0.067$) for theme and comparatively better than random for moral foundation ($0.143$), indicating that our noisy labeling approach has acceptable quality.
\begin{table*} 
\centering
    \resizebox{\textwidth}{!}{%
    \begin{tabular}{|l|l|}
      \hline
      \thead{\textbf{Themes}} & \thead{\textbf{Phrases}} \\
      \hline
      \textbf{EncourageVaccination} &  \makecell[l]{"Get your vaccine",
    "Get the shot",
    "Get jabbed",  
    "Protect our community by getting vaccinated", 
    "We can all do our part", 
    "Your shot matters", \\
    "I encourage everyone to get their shot when they can",
    "Getting vaccinated is the best way to protect yourself and everyone around you", \\
    "Take your best shot", 
    "Be the part of the solution", 
    "Win lottery by getting vaccinated",
    "Sleeve up", 
    "Vaccines save lives", 
    "Collect your vaccine incentives", \\
    "Take the vaccine for your family, for your friends, for your country",  
    "Get vaccinated", 
    "Get free vaccine and free lunch", 
    "Get boosted",  
    "Vax up",\\
    "Call us and we will answer your questions regarding COVID vaccine", 
    "Do your part to stop pandemic",
    "This is our best shot",
    "Join a discussion about vaccine"}    \\
      \hline
      \textbf{VaccineMandate} &  \makecell[l]{"Forcing people to take experimental vaccines is oppression", 
    "The vaccine has nothing to do with COVID-19, it's about the vaccine passport and tyranny",  \\
    "The vaccine mandate is unconstitutional", 
    "I'm not against the vaccine but I am against the mandate", 
    "I have freedom to choose not to take the vaccine", \\
    "I choose not to take the vaccine", 
    "I am free to refuse the vaccine",
    "It is not about COVID, it is about control", 
    "My body my choice",  \\
    "Medical segregation based on vaccine mandates is discrimination", 
    "The vaccine mandate violates my rights", 
    "I support vaccine mandate", \\
    "Firing over vaccine mandates is oppression",
    "Vaccine passports are medical tyranny", 
    "I won't let the government tell me what I should do with my body",  \\
    "I won't have the government tell me what to do", 
    "The vaccine mandate is not oppression because vaccines lower hospitalizations and death rates", \\
    "The vaccine mandate is not oppression because it will help to end this pandemic", 
    "The vaccine mandate will help us end the pandemic",  \\
    "If you don't get the vaccine based on your freedom of choice, don’t come crawling to the emergency room when you get COVID", \\
    "If you refuse a free FDA-approved vaccine for non-medical reasons, then the government shouldn't continue to give you free COVID tests",  \\
    "You are free not to take the vaccine, businesses are also free to deny you entry",
    "We need a vaccine mandate to end this pandemic", \\
    "You are free not to take the vaccine, businesses are free to protect their customers and employees",
    "Support vaccine passport",
    \\
    "If you choose not to take the vaccine, you have to deal with the consequences",  
    "We don't support vaccine passport", \\
    "If it is your body your choice, then insurance companies should stop paying for your hospitalization costs for COVID", \\
    "Check vaccine card for events", 
    "Airlines require vaccine passport", 
    "Proof of vaccine is required", 
    "Vaccine passport is useful for reopen", \\
    "Workers have declined COVID-19 Vaccine",
    "Falsely labeling the injection as a vaccine is illegal"}    \\
      \hline
     \textbf{VaccineEquity} &  \makecell[l]{"Vaccine should be available for everyone",
    "Vaccine should be free of cost", 
    "Everyone has right to get free COVID vaccine", \\
    "We don't have equal access to vaccine", 
    "We should ensure vaccine access to vulnerable communities "}    \\
      \hline
      \textbf{VaccineEfficacy} &  \makecell[l]{
     "The vaccine works",
    "The vaccine is safe",
    "Vaccines do work, ask a doctor or consult with an expert", 
    "The COVID vaccine helps to stop the spread", \\
    "Unvaccinated people are dying at a rapid rate from COVID-19", 
    "There is a lot of research supporting that vaccines work", \\
    "The research on the COVID vaccine has been going on for a long time", 
    "Millions have been vaccinated with only mild side effects", \\
    "Millions have been safely vaccinated against COVID", 
    "The benefits of the vaccine outweigh its risks", 
    "Vaccine is safe for pregnant woman", \\
    "The vaccine has benefits", 
    "The vaccine is safe for women and kids",
    "The vaccine won't make you sick",
    "The vaccine isn't dangerous", \\
    "The vaccine won't kill you", 
    "The COVID vaccine isn't a death jab",
    "The COVID vaccine doesn't harm women and kids",
    "COVID-19 vaccine ingredients are safe"}    \\
     \hline
     \textbf{GovDistrust} &  \makecell[l]{
    "We have lack of trust in the government",
    "The government is a total failure",
    "Never trust the government", 
    "Biden is a failure",
    "Biden lied people die", \\
    "The government and Fauci have been dishonest", 
    "The government always lies",
    "The government has a strong record of screwing things up", \\
    "The government is good at screwing things up",
    "The government is screwing things up", 
    "The government is lying",
    "The government only cares about money", \\
    "The government doesn't work logically",
    "Do not trust the government", 
    "The government doesn't care about people’s health", \\
    "The government won't tell you the truth about the vaccine",
    "Biden will not hold China accountable"}    \\
     \hline
     \textbf{GovTrust} &  \makecell[l]{
    "We trust the government",
    "Biden is tackling COVID",
    "The government cares for people", 
    "We are thankful to the government for the vaccine availability", \\
    "Hats off to the government for tackling the pandemic", 
    "It is a good thing to be skeptical of the government, but they are right about the COVID vaccine", \\
    "It is a good thing to be skeptical of the government, but they haven’t lied about the COVID vaccine", 
    "Biden is helping to end the pandemic",\\
    "The government can be corrupt, but they are telling the truth about the COVID vaccine", "Trump initiated COVID vaccine", \\
    "The government can be corrupt, but they are not lying about the COVID vaccine", 
    "Biden will hold China accountable"}    \\
     \hline
     \textbf{VaccineRollout} &  \makecell[l]{
    "Vaccine appointment is available",
    "Schedule your vaccine appointment", 
    "No appointment needed",
    "Walk-in vaccine clinic is available", \\
    "Drive through vaccine site is available",
    "Mobile clinic is available here", 
    "Vaccine clinic has been set up",
    "New vaccine center has been opened", \\
    "CDC recommends vaccine for kids",
    "FDA authorized COVID vaccine for children ages 5 to 11 years old"}    \\
     \hline
     \textbf{VaccineSymptom} &  \makecell[l]{
    "I got fever after taking the vaccine",
    "Know the vaccine symptom", 
    "COVID vaccines can cause blood clots", 
    "The vaccine has side effects", \\
    "The vaccine is dangerous for people with medical conditions",
    "I won't take the vaccine due to medical reasons"}    \\
     \hline
     \textbf{VaccineStatus} &  \makecell[l]{
    "Half of the population are fully vaccinated",
    "Here is the vaccine statistics",
    "Update of vaccine status", 
    "COVID-19 update", 
    "COVID death is real",\\
    "Sign up fod vaccine update",
    "We need volunteers for vaccine site",
    "Vaccination rate is slow", 
    "Vaccination rate is high", 
    "Infection rate is lower", \\
    "COVID vaccine information is here", 
    "The pandemic is not a lie, hospitalizations are on the rise"}    \\
     \hline
    \textbf{VaccineReligion} &  \makecell[l]{
    "The vaccine is against religion",
    "The vaccines are the mark of the beast",
    "The vaccine is a tool of Satan", \\
    "The vaccine is haram"
    "The vaccine is not halal", 
    "I will protect my body from a man made vaccine",
    "I put it all in God's hands", \\
    "God will decide our fate", 
    "Allah will protect us",
    "The vaccine contains bovine, which conflicts with my religion", \\
    "The vaccine contains aborted fetal tissue which is against my religion",
    "The vaccine contains pork, muslims can't take the vaccine", \\
    "Jesus will protect me",
    "The vaccine doesn't protect you from getting or spreading COVID, God does", \\
    "The COVID vaccine is another religion",
    "The vaccine is not against religion, get the vaccine", 
    "No religion ask members to refuse the vaccine", \\
    "Religious exemptions are bogus", 
    "When turning in your religious exemption forms for the vaccine, remember ignorance is not a religion", \\
    "Disregard for others' lives isn't part of your religion",  
    "Jesus is trying to protect us from COVID by divinely inspiring scientists to create vaccines"}    \\
     \hline
     \textbf{VaccineDevelopment} &  \makecell[l]{
    "COVID vaccine research has been going on for a while",
    "Plenty of research has been done on the COVID vaccine", \\
    "The technologies used to develop the COVID-19 vaccines have been in development for years to prepare for outbreaks of infectious viruses", \\
    "The testing processes for the vaccines were thorough didn't skip any steps",
    "The vaccine received FDA approval", \\
    "Vaccine uses mRNA technology",
    "the vaccine is not properly tested, it has been developed too quickly", \\
    "COVID-19 vaccines have not been through the same rigorous testing as other vaccines",
    "The COVID vaccine is experimental", \\
    "The COVID vaccine was rushed through trials",
    "The approval of the experimental vaccine was rushed"}    \\
     \hline
    \textbf{CovidPlan} &  \makecell[l]{
    "Expand vaccine distribution",
   "We are working on vaccine distribution",
   "COVID rescue plan", 
   "Cash relief",
   "More COVID testing center", \\
   "Seting up vaccine clinic", 
   "Support small business", 
   "Reopen country",
   "Reopen school",
   "Rebuilding economy", \\
   "COVID stimulus check",
   "Unemployment benefit",
   "Expand mask and PPE supply" }    \\
     \hline
    \textbf{VaccineMisinformation} &  \makecell[l]{
    "Animal shelters are empty because Dr Fauci allowed experimenting of various COVID vaccines/drugs on dogs and other domestic pets", \\
    "Fauci tortures dogs and puppies",
    "The COVID vaccine is a ploy to microchip people", 
    "Don't trust vaccine conspiracy", \\
    "Bill Gates wants to use vaccines to implant microchips in people",
    "Globalists support a covert mass chip implantation through the COVID vaccine", \\
    "There is aborted fetal tissue in the COVID Vaccines",
    "COVID vaccines contain aborted fetal cells",
    "The COVID vaccine will make you sterile", \\
    "COVID vaccine will affect your fertility",
    "The vaccine will not make you sterile",
    "The COVID vaccine will not affect your fertility", \\
    "No difference if fertility rate has been found between vaccinated and unvaccinated people", \\
    "Vaccines were tested on fetal tissues, but do not contain fetal cells", 
    "Vaccines do not contain aborted fetal cells", \\
    "Vaccine misinformation is floating around",
    "Don't believe in vaccine misinformation" }    \\
     \hline
     \textbf{NaturalImmunity} &  \makecell[l]{
    "Natural methods of protection against the disease are better than vaccines",
    "Herd immunity is broad, protective, and durable", \\
    "Natural immunity has higher level of protection than the vaccine",
    "Embrace population immunity", \\
    "I trust my immune system",
    "I have antibodies I do not need the vaccine",
    "Natural immunity is effective", \\
    "Natural immunity would require a lot of people getting sick",
    "Experts recommend the vaccine over natural immunity", \\
    "The vaccine has better long term protection than to natural immunity",
    "Natural immunity is not effective", \\
    "Experts aren’t sure how long hybrid immunity lasts",
    "Natural immunity is highly variable" }    \\
     \hline
     \textbf{Vote} &  \makecell[l]{
     "Please vote",
    "Your vote matters",
    "Go vote",
    "Vote today" }    \\
     \hline
    \end{tabular}}
    \caption{Themes and phrases to show how sponsors use social media to influence the narrative on public health crisis. Best viewed in electronic format (zoomed in).}
\label{tab:thm_phrs}
\end{table*}
\subsection{Learning Strategies}
We devise two learning strategies so that model can have access to larger datasets, which may benefit its generalization capabilities at both tasks. 
\subsubsection{Hybrid Learning}
To avoid the risk of highly biased model, instead of using fully weakly supervised batches to train the multi-task model, we create mixed batches with part gold, part noisy labels. 
\subsubsection{Two-stage Learning}
We separate the learning process into two stages, i.e., (1) pre-training stage using a large but noisy dataset, (2) fine-tuning stage using gold labels. We use a transfer learning technique by freezing the hidden layers except for the task-specific layers of our model. Therefore, we learn the model’s parameters on the weakly labeled dataset. Then, we fine-tune with fully supervised batches, i.e., gold label batches (mentioned as \textit{Two-stage learning1}), as well as with a hybrid version, i.e., involving both noisy and gold labels in the batch (named as \textit{Two-stage learning2}).
\begin{figure*}[htbp]
  \centering  
  \includegraphics[width= 1\textwidth]{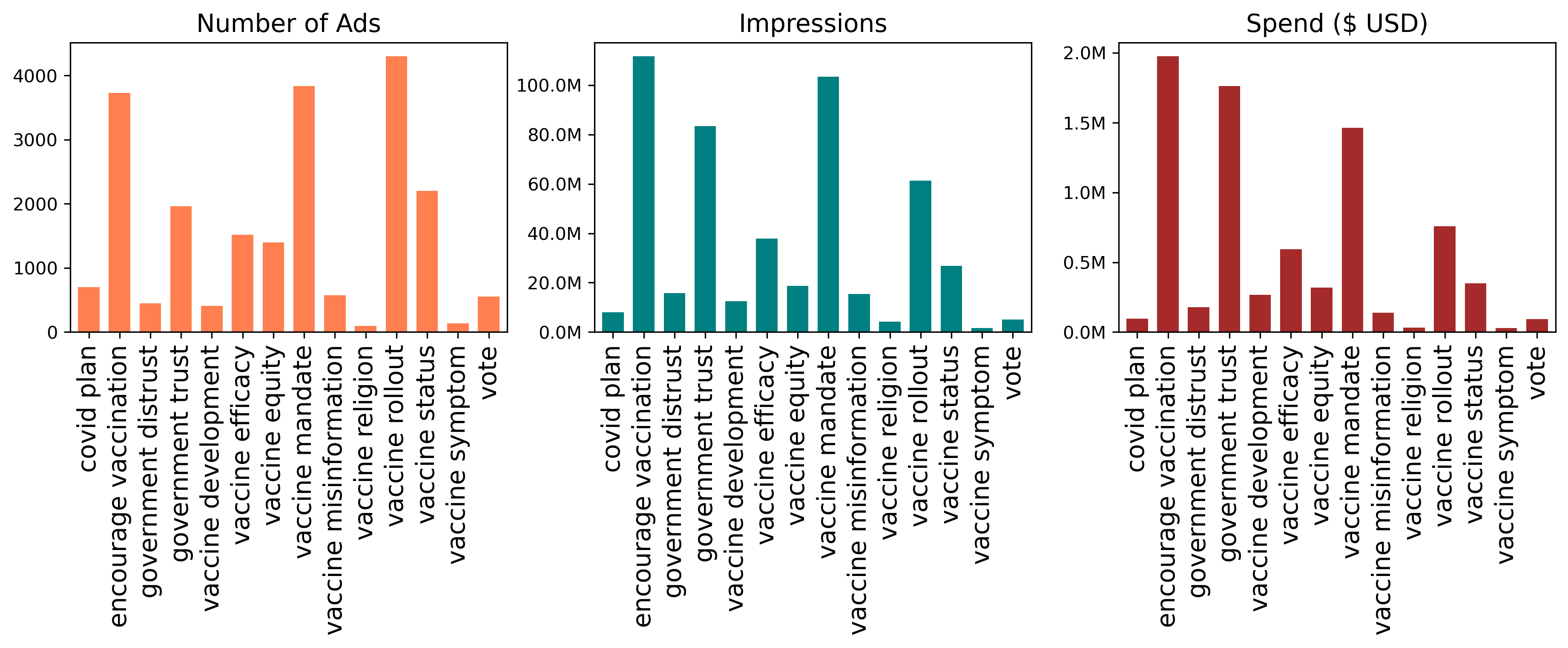}
\caption{Distribution of ad themes by number of ads, impressions, and spend.}
    \label{fig:ais}
\end{figure*}
\begin{figure*}
\begin{subfigure}{.5\columnwidth}
  \centering
  \includegraphics[width=\textwidth]{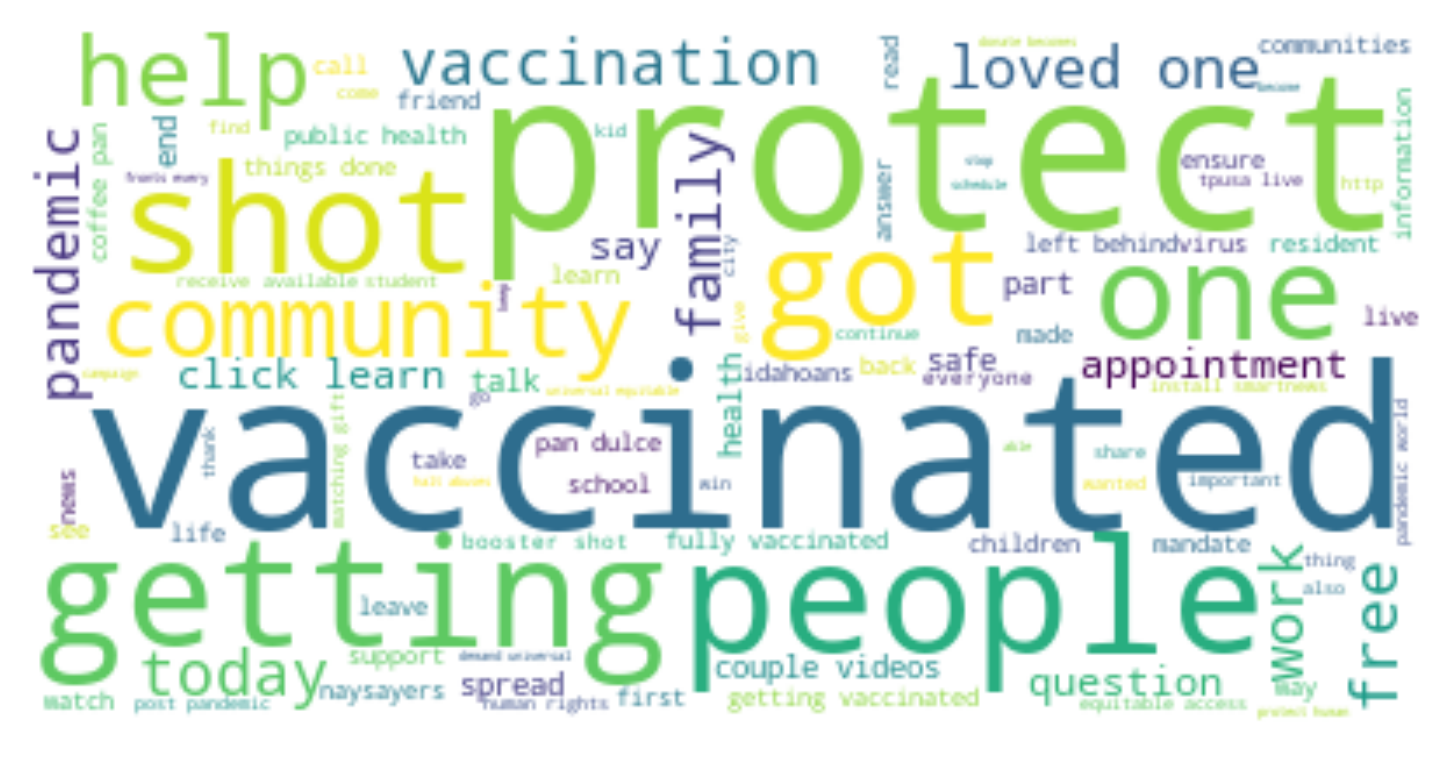}
  \caption{}
  \label{fig:envax}
\end{subfigure}%
\begin{subfigure}{.5\columnwidth}
  \centering
  \includegraphics[width=\textwidth]{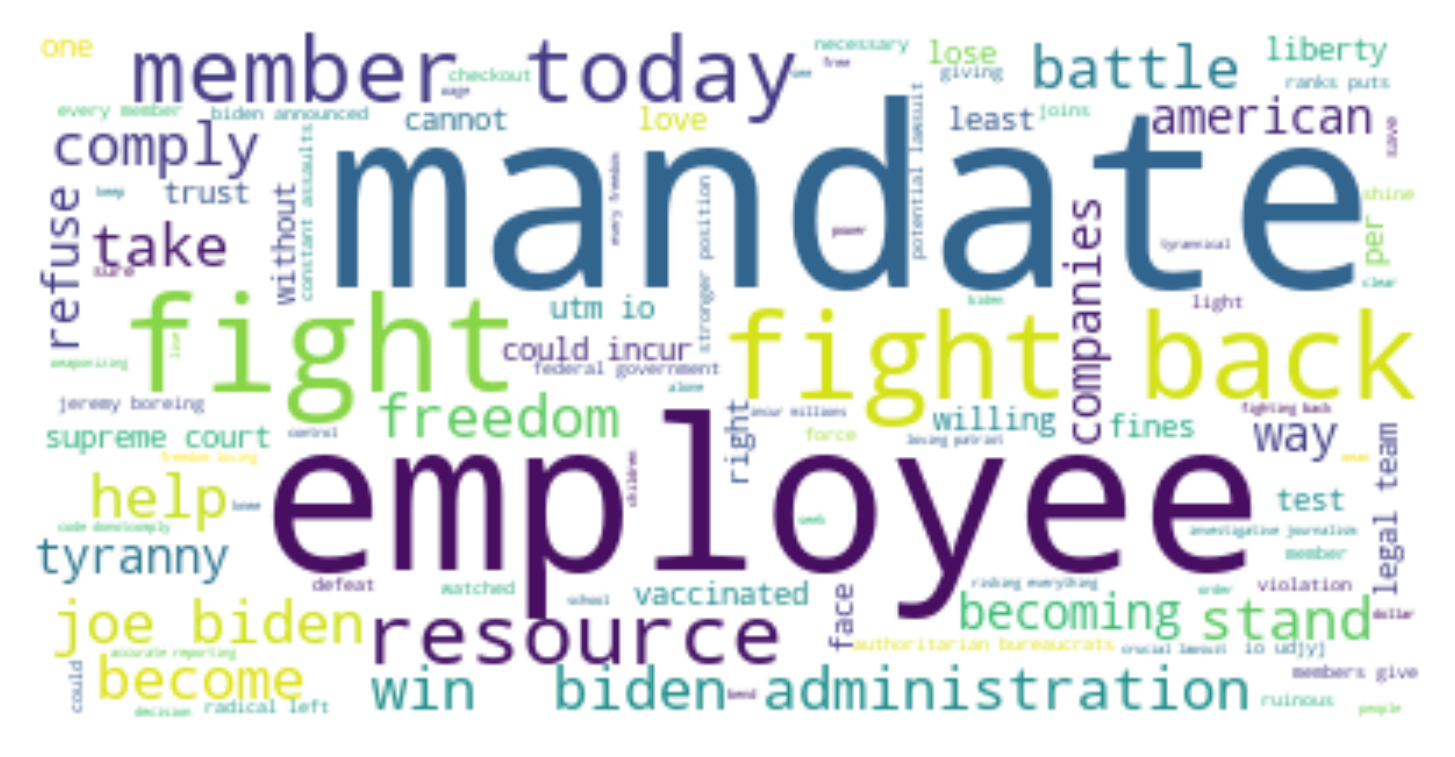}
  \caption{}
  \label{fig:vaxman}
\end{subfigure}%
\begin{subfigure}{.5\columnwidth}
  \centering
  \includegraphics[width=\textwidth]{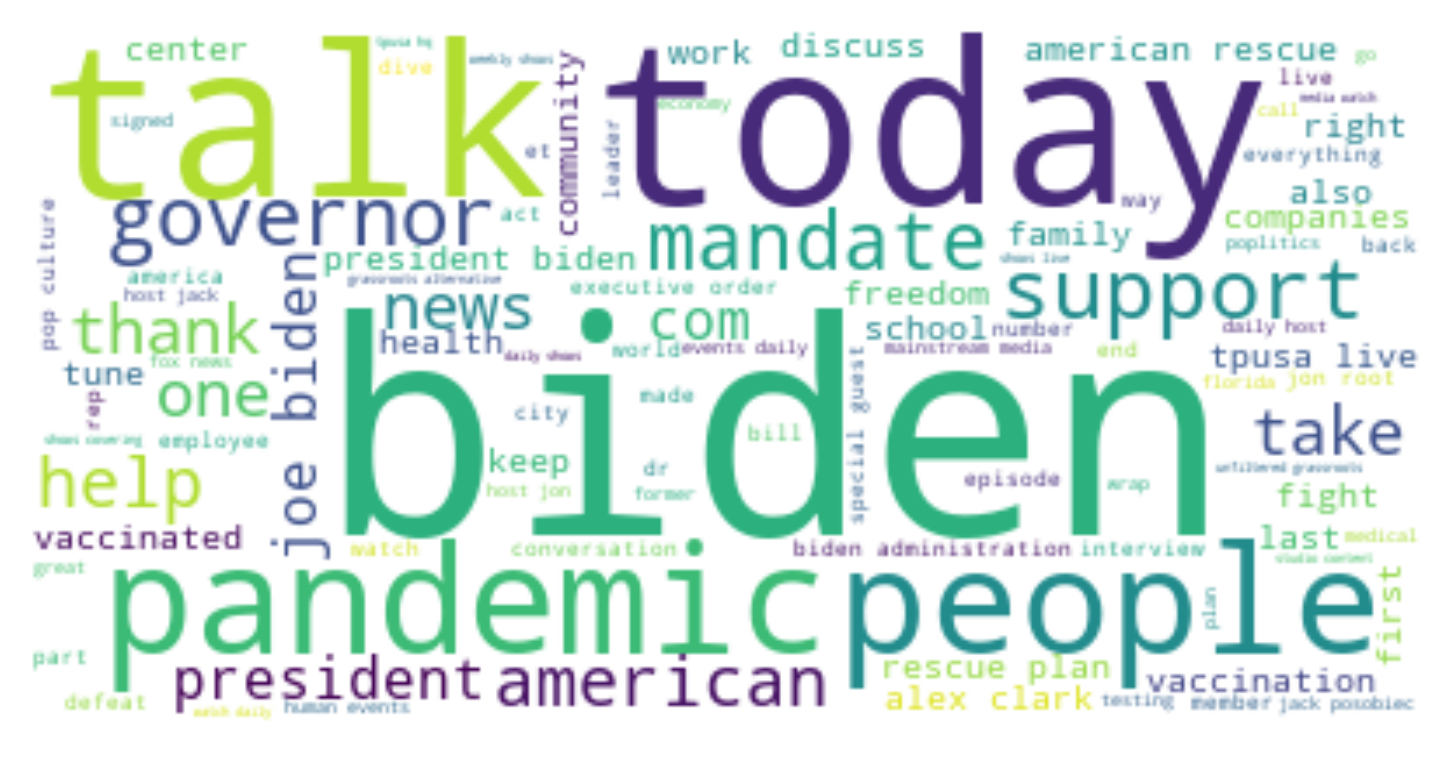}
  \caption{}
  \label{fig:govtrust}
\end{subfigure}%
\begin{subfigure}{.5\columnwidth}
  \centering
  \includegraphics[width=\textwidth]{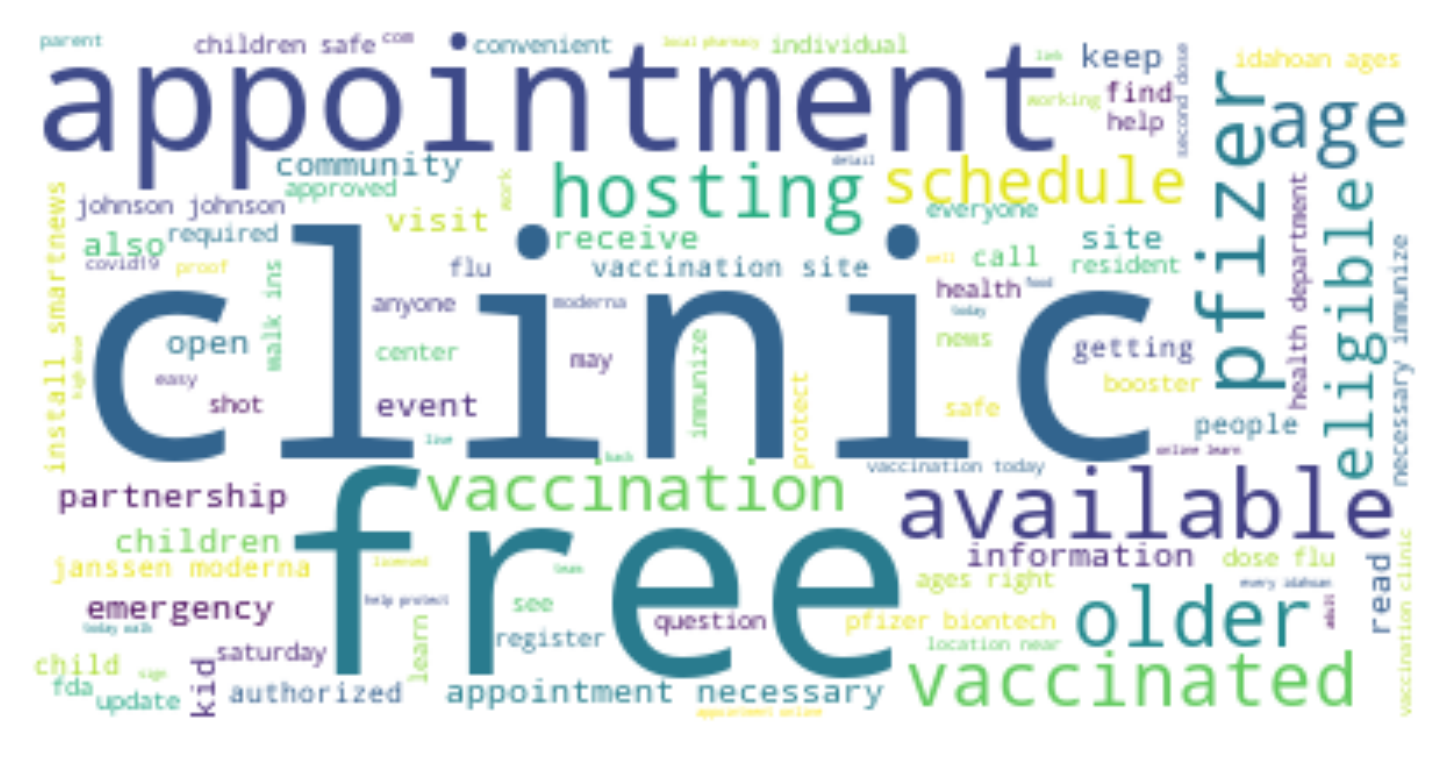}
  \caption{}
  \label{fig:vaxroll}
\end{subfigure}
\caption{Wordcloud for four messaging themes based on the popularity of ad impressions, expenditure, and number of sponsored ads. (a) encourage vaccination, (b) vaccine mandate, (c) government trust, (d) vaccine rollout. }
\label{fig:thm_wc}
\end{figure*}
\subsection{Multi-task Learning Framework}
Our proposed minimally supervised multi-task learning model consists of three main components: the text encoder, the decoder to decode the output of the encoder, 
and task specific output by summing cross-entropy losses for both tasks.
\subsubsection{Text Encoder}
We adopt a bidirectional long short-term memory network (Bi-LSTM) \cite{hochreiter1997long,schuster1997bidirectional} as our text encoder. To encode the input sentence, we first embed each word in a sentence to a low-dimensional vector
space \cite{bengio2000neural}. The input to the text encoder component is a fixed-length word sequence from an ad $w = \{w1, w2,...,w_n\}$, where $n$ is the sequence length. We use pre-trained BERT \cite{devlin2019bert} to encode the text sequence into a sequence of embeddings, $e^w = \{e^w_1,e^w_2,...,e^w_n\}$, where $e^w \in \mathbb{R}^{n \times d_e}$ and $d_e$ is the dimension of the word embedding. Bi-LSTM is employed to attain contextualized representations of words, $H = \{ h_i | h_i \in  \mathbb{R}^{2d_h \times d_e} \}$ by following operation:
\begin{align}
    h_i = [\overrightarrow {LSTM} (e_w) \oplus \overleftarrow {LSTM} (e_w)]
\end{align}
where $d_h$ denotes the dimension of hidden state from an unidirectional LSTM, while $\overrightarrow {LSTM} (.)$ and $\overleftarrow {LSTM} (.)$ stand for forward and backward LSTM, respectively and $\oplus$ represents concatenation.
\subsubsection{Decoder}
We then extract theme and MF-specific
features from the encoded hidden states, by applying linear layers and nonlinear functions, i.e., ReLU \cite{nair2010rectified}. The computation process is formulated as below:
\begin{align}
    r^t_i & = ReLU(W^t_r h_i + b^t_r)\\
    r^m_i & = ReLU(W^m_r h_i + b^m_r)
\end{align}
where $r^t_i \in \mathbb{R}^{d_r}$ and $r^m_i \in \mathbb{R}^{d_r}$ are theme and moral foundation representations, $d_r$ is the dimension of the representation. $W^t_r, W^m_r \in \mathbb{R}^{d_r \times 2d_h}$ and $b^t_r, b^m _r \in \mathbb{R}^{d_r}$ are learnable weights and biases.
\subsubsection{Task Specific Output}
Finally, we receive two series of task specific distributions by following:
\begin{align}
    P^t_i & = softmax(W^t_o r^t_i + b^t_o)\\
    P^m_i & = softmax(W^m_o r^m_i + b^m_o)
\end{align}
where  $W^t_o, W^m_o \in \mathbb{R}^{ d_l \times d_r}$ and $b^t_o, b^m _o \in \mathbb{R}^{d_l}$ are trainable parameters and $d_l$ denotes the task specific output label dimension.
We use dropout \cite{srivastava2014dropout} between individual neural network layers. Shuffled mini-batches with Adam \cite{kingma2015adam} optimizer is used. We use cross-entropy loss as the objective function as following:
\begin{align}
    \mathcal{L}_t & = - \frac{1}{n}\sum_i \hat{P}^t_i \log (P^t_i) \\
    \mathcal{L}_m & = - \frac{1}{n}\sum_i \hat{P}^m_i \log (P^m_i) 
\end{align}
where $\hat{P}^t_i$ and $\hat{P}^m_i$ represent ground truth theme and moral foundation distribution for the text.
Overall learning objective is conducted by joint training of the multi-task learning framework with the following objective:
\begin{align}
    \mathop{min}_{\theta}\mathcal{L} & = \mathcal{L}_t + \mathcal{L}_m + \gamma||\theta||_2
\end{align}
where $\theta$ stands for trainable parameters, $||\theta||_2$ is $L2$ regularization of $\theta$ and $\gamma$ is a controlling term.
\section{Experiments}
\subsection{Experimental Setup}
We evaluate a total of three settings (Fig. \ref{fig:mtl}): (1) full supervision, which trains under a fully supervised low-resources setting where we only have a small fraction of gold label data ($557$ ads) for both tasks. We randomly split the data to three subsets, namely training set ($60\%$), validation set ($20\%$), and test set ($20\%$); (2) hybrid learning experiment, at first we extract $30\%$ gold label data for mixing with weakly labeled data and keep the rest of the gold data for testing. After the mixture of gold and noisy labeled data, we randomly split it into $80\%$ training and $20\%$ validation set for hybrid learning strategy; and
(3) two-stage learning by pre-training a base model using  weak labels and then refining using both gold and hybrid labels. For pre-training, we randomly split weakly labeled data into  training ($80\%$) and validation ($20\%$) set. In fine-tuning step for \textit{two-stage learning1}, we randomly split gold labeled data train-validation-test ($60-20-20$). For \textit{two-stage learning2}, we randomly select $30\%$ of weakly labeled data and mix them with gold trained data. Then, retrain our pre-trained model by freezing the hidden layers and tweaking the task specific layers. 
\subsubsection*{Hyperparameter Setup}
In the text encoder, we set the text sequence
length, $n = 100$, and the word embedding dimension $d_e$, hidden states $d_h$, theme and MF representations $d_r$ are set to $768, 256, 128$, dropout rate = $0.2$,
batch size = $32$, learning rate = $0.001$, number of epochs = $100$. Our early stopping criterion is the lowest validation loss and we stop the learning if the loss does not decrease for $10$ consecutive epochs.

\begin{table}
\centering
\resizebox{1\columnwidth}{!}{%
\begin{tabular}{lcccc}
    \toprule
    \multirow{2}{*}{\textbf{Model}} & \multicolumn{2}{c}{\textbf{MF}} & \multicolumn{2}{c}{\textbf{Theme}} \\
    \cmidrule(r){2-3}\cmidrule(l){4-5}
     & \textbf{Accuracy}  & \textbf{Macro-avg F1}  & \textbf{Accuracy}  & \textbf{Macro-avg F1}  \\
     \midrule
     Fully supervised  & 0.688 & 0.204 & 0.250  & 0.050 \\
    \textbf{Hybrid learning} &  \textbf{0.752} & \textbf{0.510 }&  \textbf{0.690}  & \textbf{0.579} \\
    Two-stage learning1  & 0.631  & 0.202 & 0.338 & 0.135 \\
    Two-stage learning2 &  0.697 & 0.242 & 0.259 & 0.260 \\
    \bottomrule
    \end{tabular}}%
\caption{Model performance for MF and theme prediction. }
\label{tab:result}
\end{table}

\subsection{Results}
\label{Results}
We repeat each experiment $3$ times and report the average performance (with $p-value < 0.05$ by paired t-test \cite{student1908probable}) in Table \ref{tab:result}. We notice that our fully supervised baseline model struggles to predict ad themes. On the other hand, model achieves the best performance in hybrid learning strategies where training batches include both noisy and gold labels. We obtain accuracy of $75.2\%$ and macro-average F1 score of $51.0\%$ for moral foundation prediction task. In theme prediction task, we achieve $69.0\%$ accuracy and $57.9\%$ macro-average F1 score. Though both of the two-stage learning strategies obtain comparable performance in MF prediction, we see lower accuracy on theme prediction. Because of the highly imbalanced data, we notice sharp drop of macro-avg F1 score.
In conclusion, comparing this strategy to the fully supervised approach (baseline), the hybrid learning strategy seems to be a strong proposal to identify ad theme and moral foundation 
while using a multi-task learning approach. 
\subsection{Narrative Analysis}
\label{narrative}
To answer \textbf{RQ1}, we analyze the messaging strategies used by the advertisers (Fig. \ref{fig:ais}). By impressions and spend, the most popular messaging theme is `encourage vaccination', accounting for approximately $24\%$ of total spend and $22\%$ of total impressions. Narratives belonging to this category promote vaccination \textit{to protect their loved one, family, friends, and community as well as end the pandemic} using `loyalty/betrayal' moral foundation.   
Based on impression, the next most popular messaging category is `vaccine mandate', which features narratives focusing \textit{Biden's vaccine mandate respective to freedom and tyranny.}  
Based on spend, the second most popular messaging theme is `government trust', which focuses on narratives supporting \textit{Government's vaccination policy}, emphasizing `care/harm' moral foundation.
On the other hand, sponsored ads mostly have a `vaccine rollout' messaging theme ($19.7\%$) focusing on \textit{appointment availability of the free vaccine, vaccine eligibility information from FDA and CDC based on age, and information regarding drive-through/walk-in/mobile vaccine clinic.}
To show the noticeable qualitative differences, we create wordcloud with the most frequent words from those selected four messaging themes. Fig. \ref{fig:thm_wc} shows the visual representation of the text for each of $4$ theme category. Words are usually single words, and the importance of each tag is shown with font size and color.
\begin{table}
    \centering
    \resizebox{1\columnwidth}{!}{%
    \begin{tabular}{|l|l|}
      \hline
      \textbf{Type} & \textbf{Entity} \\
      \hline
      \textbf{Public health} & Children's Health System of Texas \\
      \textbf{Public health} & New York City Department of Health and Mental Hygiene \\
      \textbf{Public health} & South Carolina Department of Health \& Environmental Control \\
      \textbf{Public health} & South Dakota Department of Health \\
      \textbf{Public health} & Washington State Department of Health \\
      \textbf{Commercial} & Pfizer Inc. \\
      \textbf{Commercial} & ATTN: INC. \\
      \textbf{Commercial} & Daily Wire \\
      \textbf{Commercial} & BMO Harris Bank \\
      \textbf{Commercial} & NEWSMAX MEDIA, INC. \\
      \textbf{Political} & JB for Governor \\
      \textbf{Political} & Kemp for Governor Inc \\
      \textbf{Political} & Save America Joint Fundraising Committee \\
      \textbf{Political} & Future Majority, Inc \\
      \textbf{Political} & Terry for Virginia \\
      \textbf{Nonprofit} & Turning Point USA, Inc. \\
      \textbf{Nonprofit} & American Health Care Association and National Center for Assisted Living \\
      \textbf{Nonprofit} & PICO California Action Fund \\
      \textbf{Nonprofit} & Ad Council \\
      \textbf{Nonprofit} & PROJECT HOPE \\
    \hline
    \end{tabular}}
    \caption{List of entities.}
    \label{tab:entity}
\end{table}
\begin{table}
    \centering
    \resizebox{1\columnwidth}{!}{%
    \begin{tabular}{|l|l|}
      \hline
      \textbf{Liberal} & \textbf{Conservative} \\
      \hline
      Friends for Kathy Hochul & North Carolina Republican Party \\
      House Majority Forward & TEXANS FOR SENATOR JOHN CORNYN INC. \\
      INDIVISIBLE ACTION & JIM JORDAN FOR CONGRESS \\
     Alexandria Ocasio-Cortez for Congress & Friends of Matt Gaetz \\
     Charlie Crist for Governor & RAND PAUL FOR US SENATE \\
     Election Fund of Steven Fulop 2021 & UNSILENCED MAJORITY \\
     JAY CHEN FOR CONGRESS & Schmitt for Senate \\
     Elicker 2021 & Dr Scott Jensen for Governor \\
    \hline
    \end{tabular}}
    \caption{List of funding entities based on political view.}
    \label{tab:fe_view}
\end{table}
\begin{figure*}[htbp]
  \centering  
  \includegraphics[width= 1\textwidth]{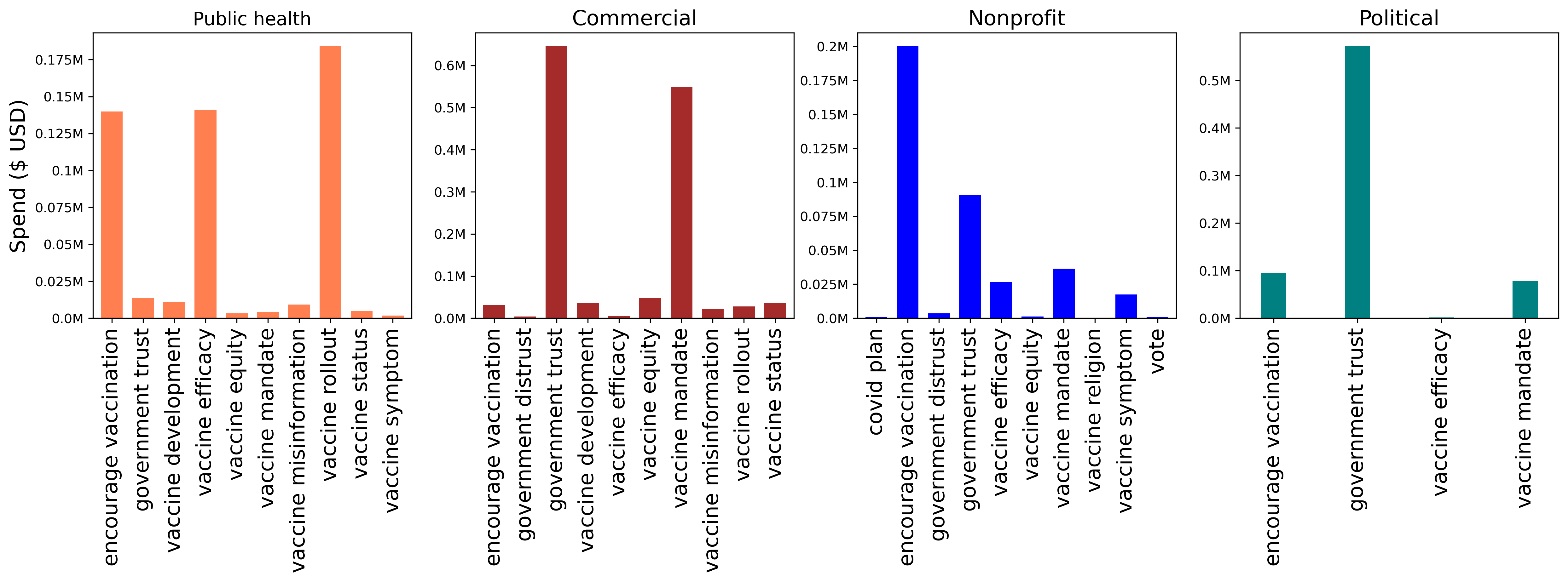}
\caption{Distribution of ad themes by funding entity type.}
    \label{fig:thm_fe}
\end{figure*}
\begin{figure*}[htbp]
  \centering  
  \includegraphics[width= 1\textwidth]{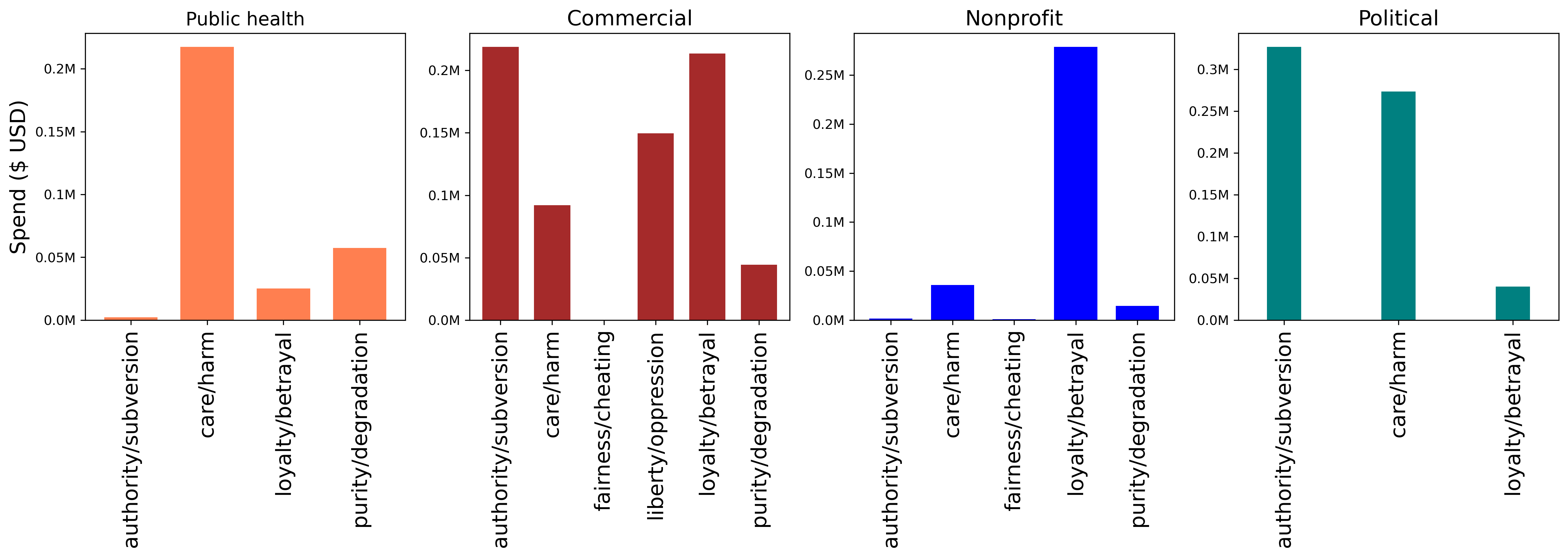}
\caption{Distribution of ad's moral foundation by funding entity type.}
    \label{fig:mf_fe}
\end{figure*}
\subsection{Distribution of Messaging by Entity Type}
\label{entity}
At first, we select top $5$ funding entities based on their expenditure.
Next, we categorize funding entities into four types (based on Facebook page category), i.e., Public health, Commercial, Nonprofit, and Political. Then, we select top $5$ funding entities based of their expenditure for each category.
Table \ref{tab:entity} shows the list of our selected entities.
Fig. \ref{fig:thm_fe} shows that the high spend on `government trust' narratives comes mostly from commercial and political entities. Public health entities spend more on `vaccine rollout', `encourage vaccination', and `vaccine efficacy' narratives. similarly, nonprofit agency focus on `encourage vaccination' theme (Fig. \ref{fig:thm_fe}). Fig. \ref{fig:mf_fe} shows the distribution of moral foundation by entity type based on expenditure. Public health entities focus more on `care/harm', where as nonprofit entities focus on `loyalty/betrayal'. Both political and commercial advertisers focus on `authority/subversion' moral foundation (Fig. \ref{fig:mf_fe}). 
\begin{figure*}[htbp]
  \centering  
  \includegraphics[width= 1\textwidth]{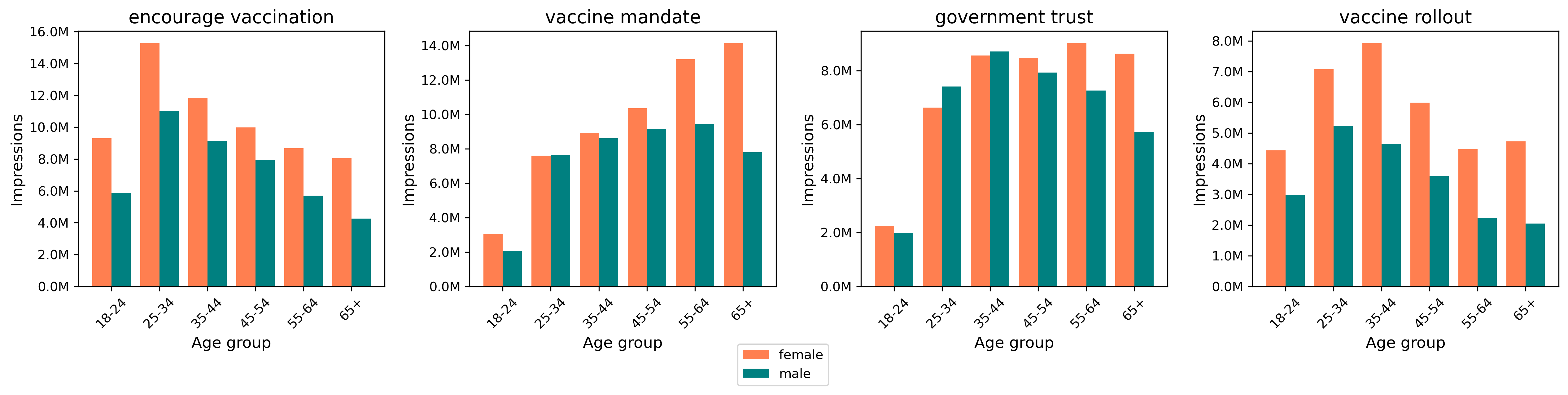}
\caption{Demographic distribution over impressions for selected four themes. T-test hypothesis testing results are in Table \ref{tab:t_test}.
}
\label{fig:demo_thm}
\end{figure*}
\begin{figure*}[htbp]
  \centering  
  \includegraphics[width= 1\textwidth]{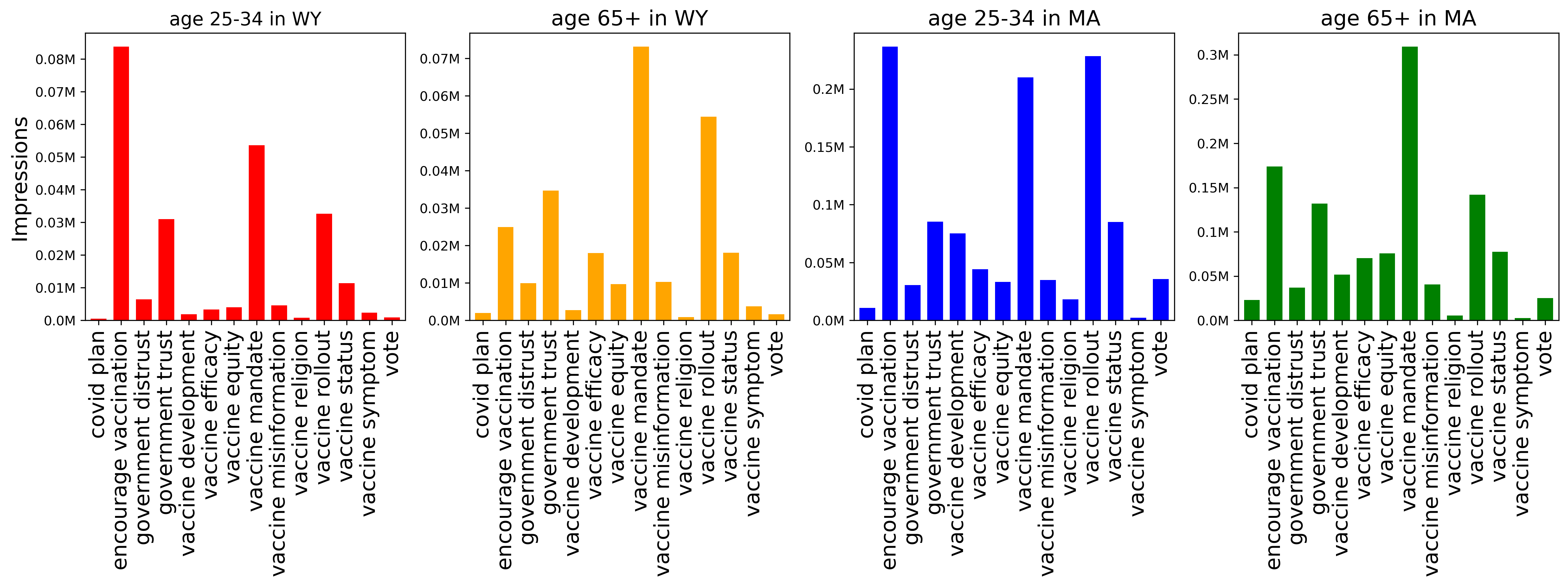}
\caption{Distribution of ad themes by demographics and geographic. T-test hypothesis testing results are in Table \ref{tab:t_test}.}
    \label{fig:thm_state}
\end{figure*}
\begin{table*}
\centering
\resizebox{1\textwidth}{!}{%
\begin{tabular}{llcc}
    \toprule
    \textbf{Null Hypothesis,} \boldmath{$H_0$} & \textbf{Alternate Hypothesis,} \boldmath{$H_a$} & \textbf{T-test statistics} & \textbf{P-value}   \\
     \midrule
     More females than the males from age range 25-34 do not view `encourage vaccination' ads.  &  More females than the males from age range 25-34 view `encourage vaccination' ads. & $9.85^{**}$ & $0.0002$  \\
     More females than males from older age (65+) do not watch `vaccine mandate' ads.  &   More females than males from older age (65+) watch `vaccine mandate' ads.   &  $2.09^{ns}$ & $0.09$  \\
     
    More females than males from older age (55+) do not watch `government trust' ads.  &   More females than males from older age (55+) watch `government trust' ads.  &  $1.37^{ns}$ & $0.229$  \\
     
     More females than the males from age range 35-44 do not view `vaccine rollout' ads.  &  More females than the males from age range 35-44 view `vaccine rollout' ads. & $8.87^{**}$ & $0.0003$  \\
     
     Population from age range ($25-34$) do not view `encourage vaccination' ads in WY.  & Population from age range ($25-34$) view `encourage vaccination' ads in WY. & $9.62^{**}$ & $0.0002$   \\
     
     Older population ($65+$) do not view narratives from `vaccine mandate' ads in WY.  &  Older population ($65+$) view narratives from `vaccine mandate' ads in WY. & $0.596^{ns}$ & $0.577$   \\
     
     Population from age range ($25-34$) do not view `encourage vaccination' ads in MA.  & Population from age range ($25-34$) view `encourage vaccination' ads in MA. & $4.02^{*}$ & $0.010$   \\
     
     Older population ($65+$) do not view narratives from `vaccine mandate' ads in MA.  &  Older population ($65+$) view narratives from `vaccine mandate' ads in MA. & $2.94^{*}$ & $0.032$   \\
    \bottomrule
    \end{tabular}}\\[.2em]
    {\small
    \resizebox{1\textwidth}{!}{%
    \begin{tabular}{l}
    	** = highly statistically significant at p-value $< 0.01$; * = statistically significant at p-value $< 0.05$; ns = statistically not significant (p-value $> 0.05$). \\
	\bottomrule
    \end{tabular}}}%
\caption{T-test of the influence of audiences' demographics and geographics on ad impressions.} 
\label{tab:t_test}
\end{table*}
To compute the statistical significance, we perform chi-square test \cite{cochran1952chi2} by taking theme distribution over moral foundation to build contingency tables separately for public health, commercial, nonprofit, and political categories. We choose the value of significance level, $\alpha = 0.05$, and our test results show that the p-value $< 0.05$ for all $4$ categories. Therefore, we reject the null hypothesis, indicating some association between theme and moral foundation. 
\begin{figure*}
\begin{subfigure}{1\textwidth}
  \centering  
  \includegraphics[width= 1\textwidth]{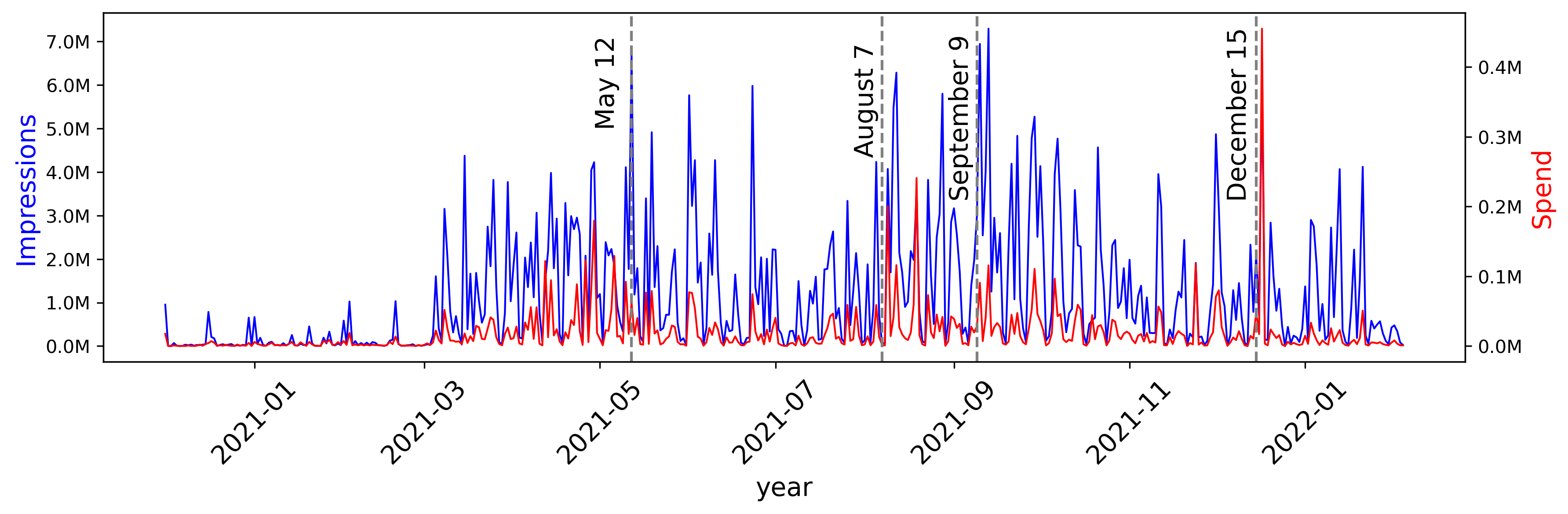}
    \caption{Timeline of Impressions and Spend}
    \label{fig:event}
\end{subfigure}
\begin{subfigure}{1\textwidth}
  \centering  
  \includegraphics[width= 1\textwidth]{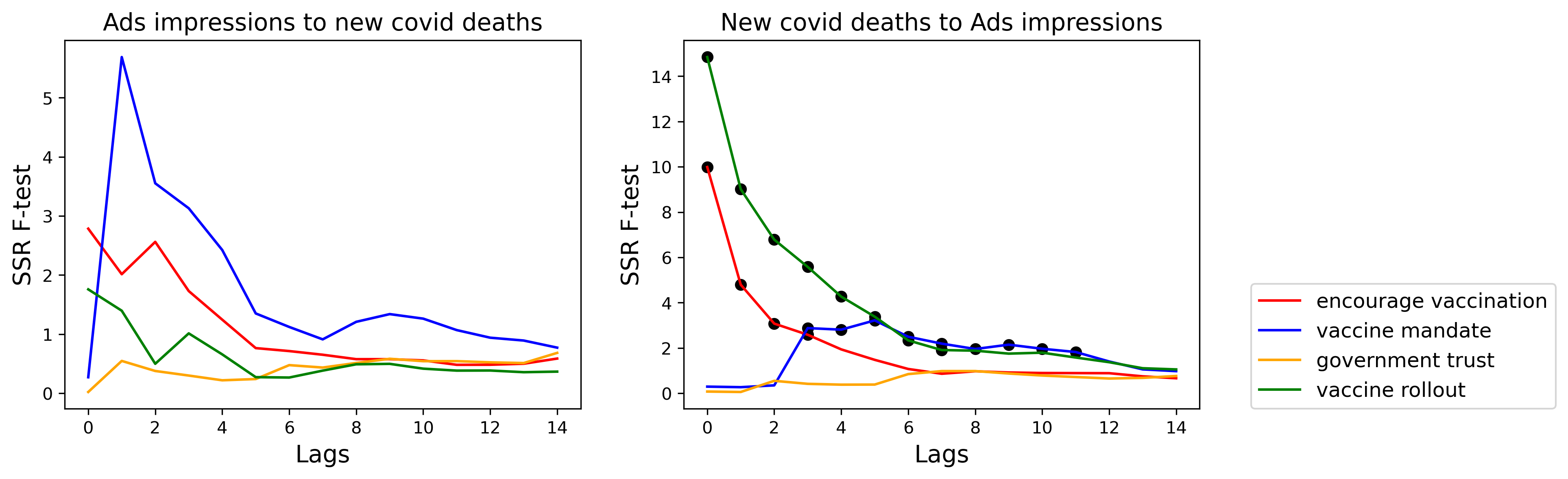}
    \caption{Granger causality test}
    \label{fig:gc}
\end{subfigure}
\caption{(a) Per day impressions and spend for ads related to COVID vaccine. We also highlight important events related to COVID with vertical bar. (b) Granger causality tests comparing daily COVID deaths and ads impressions for each day on `encourage vaccination', `vaccine mandate', `government trust', and `vaccine rollout' themes. On the $x$-axis, we report the number of lags in days; on the $y$-axis, we show the sum-of-square $F-test$ statistics for corresponding lag. We highlight those with p-value $< 0.05$ with black circle. Best viewed in electronic format (zoomed in).}
\label{fig:ts}
\end{figure*}

Looking further into the political group, we notice that this pattern primarily reflects the spending of \textit{J.B. for Governor}. This funding entity sponsored political campaign for Jay Robert (J.B.) Pritzker, who is currently serving as the $43^{rd}$ governor of Illinois. The narratives mainly focused on the `authority/subversion' moral foundation by stating \textit{how under Pritzker's leadership, Illinois administers most COVID-19 vaccine doses of any U.S. state.} Among the public health category, \textit{South Carolina Department of Health \& Environmental Control} has the highest expenditure on the `vaccine rollout' theme, which has `care/harm' moral foundation. Commercial agency \textit{Daily Wire} spends more on vaccine related ads. The \textit{Ad Council} pays the most among the nonprofit group to encourage people to get vaccinated, emphasizing on `loyalty/betrayal' moral foundation. 
\subsubsection*{Distribution of Messaging by Advertisers' Political View}
To understand existing knowledge of MFT \cite{graham2009liberals,graham2013moral}, we extend our analysis towards political advertisers' views, i.e., \textit{conservative} and \textit{liberal}. We select $16$ funding entities (Table \ref{tab:fe_view}) and look for their views at \href{https://www.opensecrets.org/outsidespending/summ.php?chrt=V&type=S}{OpenSecrets.org}. We find that liberals mostly focus on `encourage vaccination' theme and `care/harm' moral foundation whereas conservatives mainly focus on `vaccine mandate' theme and `authority/subversion' moral foundation. We again perform chi-square test by taking theme distribution over moral foundation to build contingency tables separately for liberals and conservatives funding entities. Our test results show the p-value $< 0.05$ for both views. Therefore, we reject the null hypothesis, indicating some association between theme and moral foundation.
%

After analyzing the messaging themes and moral foundations for different funding entities, the evidence indicates different groups are fulfilling different messaging roles (Answer to \textbf{RQ2}).
\subsection{Targeted Demographics and Geographics}
\label{Demographic}
In our data, ads are more prominently viewed by females than the males from all age groups based on impressions. 
However, to answer \textbf{RQ3}, we further analyze the age and gender distribution over ad impressions on four messaging themes, i.e., encourage vaccination, vaccine mandate, government trust, vaccine rollout (Fig. \ref{fig:demo_thm} ).
From Fig. \ref{fig:demo_thm} we see that more females from age group $25-34$ view `encourage vaccination' themed ads. Conversely, narratives belonging to `vaccine mandate' and `government trust' are viewed mostly by the older population. The largest gap between female and male views is found within the $ 65+$ year-old category, whereas both males and females from $25-34$ age group equally view ads having `vaccine mandate' narratives. On the other hand, more females from the age group $55-64$ watch ads with `government trust' theme and slightly more males than the females from the age group $35-44$ view those ads. We notice more females than males from the age range $35-44$ view `vaccine rollout' themed ads and there is a significant gap between female and male views in this group.
We perform t-test \cite{student1908probable} hypothesis testing to provide statistical significance of our study. Table \ref{tab:t_test} shows the null hypothesis ($H_0$), alternate hypothesis ($H_a$), t-test statistics with p-value for each tested variables. 
We select level of significance, $\alpha = 0.05$. If $ p-value < \alpha$, we reject $H_0$; otherwise we accept $H_0$. No statistical significance is found when we test whether more females compared to males of older age ($55+$) watch `vaccine mandate' and `government trust' themed ads as p-value $> 0.05$ and we accept $H_0$ ( Table \ref{tab:t_test}). We find highly statistically significant results (p-value $<0.01$) when we test whether more females than the males of the age group $25-34$ watch ads with `encourage vaccination' theme and whether more females than the males of the age group $35-44$ watch `vaccine rollout' ads. For both tests p-value $<0.01$ and we reject $H_0$ (Table \ref{tab:t_test}).

To understand how messaging changes based on geographic and demographics targeting, we choose two different age group $25-34$ and age group $65+$ from low vaccination rate state Wyoming and high vaccination rate state Massachusetts\footnote{\href{https://usafacts.org/visualizations/covid-vaccine-tracker-states/?utm_source=usnews&utm_medium=partnership&utm_campaign=fellowship&utm_content=link}{usafacts.org}}. Fig. \ref{fig:thm_state} shows that people from age range $25-34$ watch ads having `encourage vaccination' theme both in WY and MA. Our t-test results show statistical significance as p-value $<0.05$ (Table \ref{tab:t_test}). On the other hand, the older population ($65+$) views narratives from `vaccine mandate' ads both in WY and MA. But we do not find statistical significance for WY as p-value $>0.05$, and we accept the null hypothesis (Table \ref{tab:t_test}). For MA, we find statistical significance (p-value $<0.05$).
\subsection{Temporal Relationship with COVID Status}
\label{granger}
After analyzing the targeted demographics of the ads, we now move our attention to \textit{when} the audience is exposed to them. Therefore, for the temporal evolution of these campaigns and the attention audiences receive, we look at the impressions and expenditure of ads over 2021 (Fig. \ref{fig:event}). 
It is clear that there was a noticeable
spike of ad impressions on May $12^{th}$ when \href{https://www.cdc.gov/media/releases/2021/s0512-advisory-committee-signing.html}{CDC} recommended Pfizer vaccine for adolescents age 12 and older. 
There was a significant
increase in impressions on August $9^{th}$. It was after the day U.S. had seen more than $107,000$ daily cases, the highest it had seen in six months. In addition, hospitalizations were the highest since February, with most occupants unvaccinated in August $7^{th}$\footnote{\href{https://www.voanews.com/a/covid-19-pandemic_us-averaging-107000-new-covid-19-cases-day/6209310.html}{www.voanews.com}}. 
Spikes of impressions occurred after September $9^{th}$ when Biden announced new COVID-19 vaccine mandates for federal workers, large employers and health care staff\footnote{\href{https://www.whitehouse.gov/briefing-room/presidential-actions/2021/09/09/executive-order-on-requiring-coronavirus-disease-2019-vaccination-for-federal-employees/}{www.whitehouse.gov}}.
%
A significant spike for expenditure was noticed after December $15^{th}$ as COVID deaths in the United States surpassed $800,000$\footnote{\href{https://www.nytimes.com/2021/12/15/us/covid-deaths-united-states.html}{www.nytimes.com}}.
%
\subsubsection*{Granger Causality with Daily COVID Death }
As we observe large spikes, we
ask whether ads impressions are more likely to follow newly COVID death status per day. To check this, we compute two
time series:
\newline
(1) For COVID daily death data, we consider the number of new COVID death per day called $Deaths(t)$ from \href{https://data.cdc.gov/Case-Surveillance/United-States-COVID-19-Cases-and-Deaths-by-State-o/9mfq-cb36}{data.cdc.gov}.
\newline
(2) For impressions, we calculate the total number of impressions of `encourage vaccination', `vaccine mandate', `government trust', `vaccine rollout' themed ads for a given day, $AdsImpressions(t)$.

We check stationarity for both time series using Augmented Dickey-Fuller test (ADF Test) \cite{cheung1995lag}. 
As p-value $< 0.05$ for both time series, we reject the null hypothesis $(H_0)$, the data does not have a unit root and is stationary.
Then, we compute following two Granger causality tests with these two time series to check (1) Do $Deaths(t)$ \textit{Granger cause} $AdsImpressions(t)$? (2) Do $AdsImpressions(t)$ \textit{Granger cause} $Deaths(t)$? The null hypothesis $(H_0)$ assumed by the first test is that $Deaths(t)$ do not Granger cause $AdsImpressions(t)$ and the alternative hypothesis $(H_A)$ is that $Deaths(t)$ Granger cause $AdsImpressions(t)$. For the second test, $H_0$ is $AdsImpressions(t)$ do not Granger cause $Deaths(t)$ and $H_A$ = $AdsImpressions(t)$ Granger cause $Deaths(t)$. We reject $H_0$ if p-value is $< 0.05$ for these tests.
We report results for these two tests in Fig. \ref{fig:gc}. We notice a significant F-test for the hypothesis of $Deaths(t)$ Granger cause $AdsImpressions(t)$ (Right side of Fig. \ref{fig:gc}) which answers \textbf{RQ4}. Conversely, we find no significant Granger causality from ads impressions on specific theme to daily new COVID death (Left side of Fig. \ref{fig:gc}). 
Our finding from this analysis is when number of daily new COVID deaths changes, ads sponsored by the advertisers supporting specific theme get more attention.

\section{Conclusion}
We suggest a minimally supervised multi-task learning framework for analyzing COVID-19 vaccine campaigns on social media. Also, by providing a novel dataset and set of themes and phrases to analyze ongoing vaccine campaigns on social media, we hope to help policymakers make better decisions on pandemic control. Our work has some limitations, such as restricting our attention to a specific use case of Facebook advertising: COVID vaccine related ads in USA. Another is transparency -- some particular aspects of the advertising campaigns are not available to the public through the Facebook Ads Library API, thus limiting our findings. Despite these limitations, we believe our work brings essential elements to the debate on Facebook advertising in public health crisis. As we make our dataset available to the community, we hope the advertising domain will become a crucial part of public discourse on public health. This work is only the first step toward a more transparent campaign which we hope will continue. 
\section{Ethics Statement}
The data collected in the paper was made publicly available by Facebook Ads API. The data does not contain any personally identifying information. Our data reports engagement patterns at an aggregate level. Therefore, we do not derive or infer any potentially sensitive characteristics or health information that may violate users’ privacy. Our analysis is based on English-written ads focusing on the United States only, which may have unknown biases. 
\section*{Acknowledgements}
We are thankful to the anonymous reviewers for their insightful comments. This work was partially supported by Purdue Graduate School Summer Research Grant and an NSF CAREER award IIS-2048001.
\bibliography{covid_bib}{}

\begin{thebibliography}{10}

\bibitem{cucinotta2020declares}
D.~Cucinotta and M.~Vanelli, ``Who declares covid-19 a pandemic,'' {\em Acta
  Bio Medica: Atenei Parmensis}, 2020.

\bibitem{tagliabue2020pandemic}
F.~Tagliabue, L.~Galassi, and P.~Mariani, ``The “pandemic” of
  disinformation in covid-19,'' {\em SN comprehensive clinical medicine}, 2020.

\bibitem{mcleod2007maslow}
S.~McLeod, ``Maslow's hierarchy of needs,'' {\em Simply psychology}, 2007.

\bibitem{haidt2004intuitive}
J.~Haidt and C.~Joseph, ``Intuitive ethics: How innately prepared intuitions
  generate culturally variable virtues,'' {\em Daedalus}, 2004.

\bibitem{haidt2007morality}
J.~Haidt and J.~Graham, ``When morality opposes justice: Conservatives have
  moral intuitions that liberals may not recognize,'' {\em SJR}, 2007.

\bibitem{graham2009liberals}
J.~Graham, J.~Haidt, and B.~A. Nosek, ``Liberals and conservatives rely on
  different sets of moral foundations.,'' {\em JPSP}, 2009.

\bibitem{graham2013moral}
J.~Graham, J.~Haidt, {\em et~al.}, ``Moral foundations theory: The pragmatic
  validity of moral pluralism,'' in {\em Adv Exp Soc Psychol}, 2013.

\bibitem{weber2013moral}
C.~R. Weber and C.~M. Federico, ``Moral foundations and heterogeneity in
  ideological preferences,'' {\em Political Psychology}, 2013.

\bibitem{silver2017conservatives}
J.~R. Silver and E.~Silver, ``Why are conservatives more punitive than
  liberals? a moral foundations approach.,'' {\em LHB}, 2017.

\bibitem{johnson2018classification}
K.~Johnson and D.~Goldwasser, ``Classification of moral foundations in
  microblog political discourse,'' in {\em ACL}, 2018.

\bibitem{roy2021identifying}
S.~Roy, M.~L. Pacheco, and D.~Goldwasser, ``Identifying morality frames in
  political tweets using relational learning,'' in {\em EMNLP}, 2021.

\bibitem{turner2021conservatives}
F.~M. Turner-Zwinkels, B.~B. Johnson, {\em et~al.}, ``Conservatives’ moral
  foundations are more densely connected than liberals’ moral foundations,''
  {\em PSPB}, 2021.

\bibitem{wawrzuta2021arguments}
D.~Wawrzuta, M.~Jaworski, {\em et~al.}, ``What arguments against covid-19
  vaccines run on facebook in poland: Content analysis of comments,'' {\em
  Vaccines}, 2021.

\bibitem{weinzierl2021misinformation}
M.~Weinzierl, S.~Hopfer, and S.~M. Harabagiu, ``Misinformation adoption or
  rejection in the era of covid-19,'' in {\em ICWSM}, 2021.

\bibitem{pacheco2022holistic}
M.~L. Pacheco, T.~Islam, M.~Mahajan, A.~Shor, M.~Yin, L.~Ungar, and
  D.~Goldwasser, ``A holistic framework for analyzing the covid-19 vaccine
  debate,'' {\em NAACL}, 2022.

\bibitem{pagliaro2021trust}
S.~Pagliaro, S.~Sacchi, {\em et~al.}, ``Trust predicts covid-19 prescribed and
  discretionary behavioral intentions in 23 countries,'' {\em PloS one}, 2021.

\bibitem{diaz2021reactance}
R.~D{\'\i}az and F.~Cova, ``Reactance, morality, and disgust: The relationship
  between affective dispositions and compliance with official health
  recommendations during the covid-19 pandemic,'' {\em Cognition and Emotion},
  2021.

\bibitem{chan2021moral}
E.~Y. Chan, ``Moral foundations underlying behavioral compliance during the
  covid-19 pandemic,'' {\em Personality and individual differences}, 2021.

\bibitem{shurafa2020political}
C.~Shurafa, K.~Darwish, and W.~Zaghouani, ``Political framing: Us covid19 blame
  game,'' in {\em SocInfo}, 2020.

\bibitem{muric2021covid}
G.~Muric, Y.~Wu, {\em et~al.}, ``Covid-19 vaccine hesitancy on social media:
  building a public twitter data set of antivaccine content, vaccine
  misinformation, and conspiracies,'' {\em JMIR}, 2021.

\bibitem{thelwall2021covid}
M.~Thelwall, K.~Kousha, and S.~Thelwall, ``Covid-19 vaccine hesitancy on
  english-language twitter,'' {\em EPI}, 2021.

\bibitem{krawczyk2021quantifying}
K.~Krawczyk, T.~Chelkowski, , {\em et~al.}, ``Quantifying online news media
  coverage of the covid-19 pandemic: Text mining study and resource,'' {\em
  JMIR}, 2021.

\bibitem{jing2021characterizing}
E.~Jing and Y.-Y. Ahn, ``Characterizing partisan political narratives about
  covid-19 on twitter,'' {\em arXiv preprint arXiv:2103.06960}, 2021.

\bibitem{prado2022moral}
M.~F. Prado, V.~P. Bustos, {\em et~al.}, ``Moral narratives around the
  vaccination discourse on the facebook platform,'' {\em arXiv preprint
  arXiv:2206.01598}, 2022.

\bibitem{shi2021psycho}
J.~Shi, P.~Ghasiya, {\em et~al.}, ``Psycho-linguistic differences among
  competing vaccination communities on social media,'' {\em APSIPA}, 2021.

\bibitem{bokemper2022testing}
S.~E. Bokemper, G.~A. Huber, {\em et~al.}, ``Testing persuasive messaging to
  encourage covid-19 risk reduction,'' {\em PloS one}, 2022.

\bibitem{mermin2022s}
K.~Mermin-Bunnell and W.-k. Ahn, ``It’s time to be disgusting about covid-19:
  Effect of disgust priming on covid-19 public health compliance among liberals
  and conservatives,'' {\em Plos one}, 2022.

\bibitem{nan2022public}
X.~Nan, I.~A. Iles, {\em et~al.}, ``Public health messaging during the covid-19
  pandemic and beyond: Lessons from communication science,'' {\em Health
  Communication}, 2022.

\bibitem{graham2020faith}
A.~Graham, F.~T. Cullen, {\em et~al.}, ``Faith in trump, moral foundations, and
  social distancing defiance during the coronavirus pandemic,'' {\em Socius},
  2020.

\bibitem{heine2021using}
F.~Heine and E.~Wolters, ``Using moral foundations in government communication
  to reduce vaccine hesitancy,'' {\em PloS one}, 2021.

\bibitem{tarry2022political}
H.~Tarry, V.~V{\'e}zina, J.~Bailey, and L.~Lopes, ``Political orientation,
  moral foundations, and covid-19 social distancing,'' {\em PloS one}, 2022.

\bibitem{coelho2022left}
G.~L. Coelho, L.~J. Wolf, {\em et~al.}, ``Do left-wingers discriminate? a
  cross-cultural study on the links between political orientation, values,
  moral foundations, and the covid-19 passport,'' {\em PsyArXiv.}, 2022.

\bibitem{kyung2022political}
E.~Kyung, M.~Thomas, and A.~Krishna, ``How political identity influences
  covid-19 risk perception: A model of identity-based risk perception,'' {\em
  JACR}, 2022.

\bibitem{mejova2020covid}
Y.~Mejova and K.~Kalimeri, ``Covid-19 on facebook ads: competing agendas around
  a public health crisis,'' in {\em SIGCAS}, 2020.

\bibitem{silva2021covid}
M.~Silva and F.~Benevenuto, ``Covid-19 ads as political weapon,'' in {\em SAC},
  2021.

\bibitem{belem2021weakly}
C.~Bel{\'e}m, V.~Balayan, P.~Saleiro, and P.~Bizarro, ``Weakly supervised
  multi-task learning for concept-based explainability,'' {\em arXiv preprint
  arXiv:2104.12459}, 2021.

\bibitem{mekala2020meta}
D.~Mekala, X.~Zhang, and J.~Shang, ``Meta: Metadata-empowered weak supervision
  for text classification,'' in {\em EMNLP}, 2020.

\bibitem{ratner2018snorkel}
A.~Ratner, B.~Hancock, {\em et~al.}, ``Snorkel metal: Weak supervision for
  multi-task learning,'' in {\em DEEM}, 2018.

\bibitem{islam2022twitter}
T.~Islam and D.~Goldwasser, ``Twitter user representation using weakly
  supervised graph embedding,'' in {\em ICWSM}, 2022.

\bibitem{caruana1997multitask}
R.~Caruana, ``Multitask learning,'' {\em Machine learning}, 1997.

\bibitem{liu2016recurrent}
P.~Liu, X.~Qiu, and X.~Huang, ``Recurrent neural network for text
  classification with multi-task learning,'' {\em arXiv preprint
  arXiv:1605.05101}, 2016.

\bibitem{liu2019multi}
X.~Liu, P.~He, {\em et~al.}, ``Multi-task deep neural networks for natural
  language understanding,'' in {\em ACL}, 2019.

\bibitem{lu2020multi}
G.~Lu, J.~Gan, {\em et~al.}, ``Multi-task learning using a hybrid
  representation for text classification,'' {\em Neural Comput. Appl.}, 2020.

\bibitem{vandenhende2021multi}
S.~Vandenhende, S.~Georgoulis, {\em et~al.}, ``Multi-task learning for dense
  prediction tasks: A survey,'' {\em TPAMI}, 2021.

\bibitem{cohen1960coefficient}
J.~Cohen, ``A coefficient of agreement for nominal scales,'' {\em EPM}, 1960.

\bibitem{weinzierl2022hesitancy}
M.~Weinzierl and S.~Harabagiu, ``From hesitancy framings to vaccine hesitancy
  profiles: A journey of stance, ontological commitments and moral
  foundations,'' {\em arXiv preprint arXiv:2202.09456}, 2022.

\bibitem{cheng2022debate}
F.~K. Cheng, ``Debate on mandatory covid-19 vaccination,'' {\em Ethics,
  Medicine and Public Health}, 2022.

\bibitem{ehde2021covid}
D.~M. Ehde, M.~K. Roberts, {\em et~al.}, ``Covid-19 vaccine hesitancy in adults
  with multiple sclerosis in the united states: a follow up survey during the
  initial vaccine rollout in 2021,'' {\em MSARD}, 2021.

\bibitem{roozenbeek2020susceptibility}
J.~Roozenbeek, C.~R. Schneider, {\em et~al.}, ``Susceptibility to
  misinformation about covid-19 around the world,'' {\em R. Soc. Open Sci.},
  2020.

\bibitem{reimers2019sentence}
N.~Reimers and I.~Gurevych, ``Sentence-bert: Sentence embeddings using siamese
  bert-networks,'' {\em arXiv preprint arXiv:1908.10084}, 2019.

\bibitem{rousseeuw1987silhouettes}
P.~J. Rousseeuw, ``Silhouettes: a graphical aid to the interpretation and
  validation of cluster analysis,'' {\em J. Comput. Appl. Math}, 1987.

\bibitem{vanDerMaaten2008}
L.~van~der Maaten and G.~Hinton {\em JMLR}, 2008.

\bibitem{pennebaker2001linguistic}
J.~W. Pennebaker, M.~E. Francis, and R.~J. Booth, ``Linguistic inquiry and word
  count: Liwc 2001,'' {\em Mahway: Lawrence Erlbaum Associates}, 2001.

\bibitem{tausczik2010psychological}
Y.~R. Tausczik and J.~W. Pennebaker, ``The psychological meaning of words: Liwc
  and computerized text analysis methods,'' {\em JLS}, 2010.

\bibitem{hochreiter1997long}
S.~Hochreiter and J.~Schmidhuber, ``Long short-term memory,'' {\em Neural
  computation}, 1997.

\bibitem{schuster1997bidirectional}
M.~Schuster and K.~K. Paliwal, ``Bidirectional recurrent neural networks,''
  {\em IEEE Trans. Signal Process.}, 1997.

\bibitem{bengio2000neural}
Y.~Bengio, R.~Ducharme, and P.~Vincent, ``A neural probabilistic language
  model,'' {\em NeurIPS}, 2000.

\bibitem{devlin2019bert}
J.~Devlin, M.-W. Chang, {\em et~al.}, ``Bert: Pre-training of deep
  bidirectional transformers for language understanding,'' in {\em NAACL-HLT},
  2019.

\bibitem{nair2010rectified}
V.~Nair and G.~E. Hinton, ``Rectified linear units improve restricted boltzmann
  machines,'' in {\em ICML}, 2010.

\bibitem{srivastava2014dropout}
N.~Srivastava, G.~Hinton, {\em et~al.}, ``Dropout: a simple way to prevent
  neural networks from overfitting,'' {\em JMLR}, 2014.

\bibitem{kingma2015adam}
D.~P. Kingma and J.~Ba, ``Adam: A method for stochastic optimization,'' in {\em
  ICLR (Poster)}, 2015.

\bibitem{student1908probable}
Student, ``Probable error of a correlation coefficient,'' {\em Biometrika},
  1908.

\bibitem{cochran1952chi2}
W.~G. Cochran, ``The $\chi$2 test of goodness of fit,'' {\em The Annals of
  mathematical statistics}, 1952.

\bibitem{cheung1995lag}
Y.-W. Cheung and K.~S. Lai, ``Lag order and critical values of the augmented
  dickey--fuller test,'' {\em JBES}, 1995.

\end{thebibliography}
\bibliographystyle{ieeetr}
\end{document}